\documentclass[10pt,a4paper]{IEEEtran}
\usepackage{url}
\usepackage{verbatim}
\usepackage{epsfig}
\usepackage{ifthen}
\usepackage{law}
\usepackage{mystyle}
\usepackage{graphics}
\usepackage{times}
\usepackage{cite}
\usepackage{bbding}
\usepackage{amsfonts}
\usepackage{amsmath}
\usepackage{amssymb}
\usepackage{algorithm}

\usepackage{multirow}
\usepackage{hhline}
\usepackage{epstopdf}
\usepackage{subfigure}
\usepackage{amssymb}
\usepackage{amsmath}
\newcommand{\argmax}{\arg\!\max}
\newcommand{\argmin}{\arg\!\min}
\usepackage{graphicx}
\usepackage{algpseudocode}
\usepackage{picins}

\newcommand{\Draft}{0}   

\ifthenelse{\Draft = 1}
{
  \newcommand{\fixMe}[2][]{
    \typeout{***** ERROR: fixMe still in final version *****}
  }
}
{
  \newcommand{\fixMe}[2][] {[{\bf #1}] {\bf \marginpar{\large FIX}} {\em #2}}

}

\title{
Graph Regularized Non-negative Matrix Factorization By Maximizing Correntropy
\thanks{Manuscript submitted to Journal of Computers.}
\thanks{}
}

\author{
Le Li$^{1}$, Jianjun Yang$^{2}$, Kaili Zhao$^{3}$, Yang Xu$^{3}$, Honggang Zhang$^{3}$, Zhuoyi Fan$^{4}$\\
\normalsize $^{1}$ School of Computer Science, University of Waterloo, Ontario N2L3G1, Canada\\
\normalsize $^{2}$  Department of Computer Science, University of North Georgia, Oakwood, GA 30566, USA\\
\normalsize $^{3}$  PRIS Lab, Beijing University of Posts and Telecommunications, Beijing 100876, P.R.China\\
\normalsize $^{4}$  SEEE, Huazhong University of Science and Technology, Hubei 430074, P.R.China\\
Email: l248li@uwaterloo.ca, jianjun.yang@ung.edu, \{xj992adolphxy, kailizhao1989\}@gmail.com, zhhg@bupt.edu.cn, fanzhuoyi@hust.edu.cn
}

\date{}
\begin{document}

\maketitle
\thispagestyle{empty}

\begin{abstract}
Non-negative matrix factorization (NMF) has proved effective in many clustering and classification tasks. The classic ways to measure the errors between the original and the reconstructed matrix are $l_2$ distance or Kullback-Leibler (KL) divergence. However, nonlinear cases are not properly handled when we use these error measures. As a consequence, alternative measures based on nonlinear kernels, such as correntropy, are proposed. However, the current correntropy-based NMF only targets on the low-level features without considering the intrinsic geometrical distribution of data. In this paper, we propose a new NMF algorithm that preserves local invariance by adding graph regularization into the process of max-correntropy-based matrix factorization. Meanwhile, each feature can learn corresponding kernel from the data. The experiment results of Caltech101 and Caltech256 show the benefits of such combination against other NMF algorithms for the unsupervised image clustering.
\end{abstract}




\section{Introduction}
\label{sec:intro}
Given a collection of images, the clustering algorithms attempt to group the dataset into multiple clusters such that images in the same cluster are similar to each other in terms of the semantics information. In this process, a good feature extraction method is vital to the clustering performance. Essentially, clustering (e.g. \textit{k}-means) or classification algorithms (e.g. support vector machines \cite{suykens1999least,chapelle1999support,zhou2010region,li2010adaptive}) map the low-level image features to semantic information. These algorithms have a variety of applications in different areas\cite{FXR_iccv13,FT_eccv12,qin2013social,qin2012improving,yu2010adaptive,yu2011nonparametric,sun2013novel,song2011image,xu2013cross,zhang2011distributed,sun2013space,sun2013mobile,shen2013virtual,shen2013layer,wang2012scimate}. If the extracted image features can reflect the latent semantic concepts, we believe it can somehow better boost the clustering/classification performance.

Recently, non-negative matrix factorization (NMF) has proven to be a powerful matrix factorization tool for data representation. Matrix factorization decomposes the original matrix $X$ into multiple low-rank matrices, such that  their product approximates $X$. In NMF, $X$ is decomposed into two non-negative matrices $H$ (basis matrix) and $W$ (coefficient matrix). The vectors in $H$ spans a latent semantic space where each basis vector defines a semantic topic. By doing so, unlike other matrix factorization methods, such as singular value decomposition, that interpret the data as both additive and subtractive combination of semantics NMF only allows additive relationship. This constraint has proved to be closer to the way how humans perceive and understand the data \cite{palmer1977hierarchical,logothetis1996visual,wachsmuth1994recognition}. Based on this methodology, we can map the low-level image features into the additive combination of latent semantics for the clustering.

Extensive work has been done to investigate the NMF algorithm for different clustering tasks. Generally, the matrix decomposition is done by minimizing the errors between original and reconstructed matrix using $l_2$ distance or KL divergence \cite{xu2003document,shahnaz2006document,seung2001algorithms}. One of the concerns is that they are linear similarity measures, which may not be suitable for data with nonlinear structure, like images \cite{sandler2011nonnegative}. A possible solution is to use nonlinear similarity measure to model the error (e.g. kernlized nonlinear mapping \cite{xu2004document,zhang2006non}). Among these nonlinear methods, we are especially interested in the NMF based on maximizing correntropy criterion (MCC). Correntropy is a generalized nonlinear measure between two variables. MCC-based methods have proved effective in many areas, e.g. cancer clustering \cite{wang2013non}, face recognition \cite{he2011maximum} and software defect prediction\cite{chang2011software}. Another approach is preserving the geometric structure of data based on the manifold assumption of data distribution. To be more specific, the authors in \cite{cai2011graph,cai2008non} exploit the local invariance and encode such geometrical information by constructing a nearest neighbor graph. Thus, at least two factors are included in modeling this process: the distance between $X$ and $H*W$ based on $l_2$ distance (or KL divergence) at low-level feature space, and the graph regularization. Furthermore, the authors in \cite{shen2010non} incorporate such intrinsic geometric information in multiple manifolds.

In this paper, we propose a graph regularized NMF algorithm based on maximizing correntropy criterion for unsupervised image clustering. We can leverage MCC to properly model the errors in low-level feature space. Furthermore, the graph regularization can keep the correct geometric information in our factorization process. To our knowledge, this is the first work that combines MCC and graph regularization in NMF. Meanwhile, our proposed algorithm can self-learn the kernels from the data to evolve for clustering task. The results on two datasets (Caltech101 and Caltech256) shows the supremacy of our proposed algorithm over other variants of NMF algorithms.

The paper is organized into six sections: related work is discussed in Section~\ref{sec:relatedWork}. In section~\ref{sec:alg}, our proposed model and the convergence proof are explained. The experiment settings are mentioned in Section~\ref{sec:expSettings}. The results and further discussions are shown in Section~\ref{sec:results}. The last section is the conclusion.

\section{Related Work}
\label{sec:relatedWork}
\textit{K}-means is perhaps one of the most popular clusteirng algorithms. This algorithm finds the match cluster for each data point by minimizing distance between the data point and existing clusters' centroids. The clusters' centroids are updated at each iteration when a new cluster member is introduced. Similar to Naive Bayes and Gaussian mixture model \cite{baker1998distributional,liu2002document} that have their own assumption of data distribution, \textit{K}-means algorithm assumes that data within the same class should also be close to each other in the feature space. One potential concern is that if the dataset properties don't follow such assumptions, the accuracy of the algorithm may be at risk.

To solve the above issue, some methods (e.g. Latent Semantics Indexing (LSI) \cite{deerwester1990indexing}) try to map the data into the latent semantics space where each basis axis in such space represents one type of semantic information of the dataset. Thus, we represent each image as a combination of multiple semantics information. Then we can use any clustering or classification algorithm to directly work on these semantic features. In LSI, the coefficients of the combination could be either positive or negative. However, a negative coefficient could be difficult to interpret. Meanwhile, LSI algorithm requires the basis in semantic space to be orthogonal from each other. This property, on one hand, ensures that we will have a unique solution \cite{ding2006orthogonal}; on the other hand, it indicates that the semantics basis are distinguished from each other, which, however, is not always the case.

NMF is similar to LSI algorithm in the way that they both map the dataset into the latent feature space. However, the basis in latent feature space are not necessarily orthogonal from each other in NMF. Meanwhile, each basis now corresponds to one topic of the dataset. The key benefit is that we can easily find the category of the image by simply investigating the largest components in the latent space. Furthermore, every element in the two decomposed low-rank matrices are non-negative. This additive combination makes it easier to interpret an image intuitively.

NMF has been investigated heavily for image clustering \cite{xu2003document,shahnaz2006document,lin2007projected,wang2012adaptive}. The classic NMF mainly targets on minimizing the $l_2$ distance or KL divergence. One issue with it is that it may not be able to handle the nonlinear data. Different algorithms have been proposed to solve this problem. One way is to use nonlinear distance measure, such as correntropy \cite{wang2013non,he2011maximum}, kernelized distance \cite{zhang2006non}, etc. Another approach is preserving the geometric structure of data based on the manifold assumption. For example, the authors in \cite{cai2011graph} exploit the local invariance and encode such geometrical information by constructing a nearest neighbor graph. In a similar way, the authors in \cite{xiao2013class,liu2012constrained} incorporate a portion of true labels into the graph-regularized method (i.e. semi-supervised learning). The key ideas are to map samples with same ground-truth labels onto the same point in the semantic feature space. Some other solutions can be found in \cite{sun2012unsupervised}.

\section{Algorithms}
\label{sec:alg}
\subsection{NMF algorithm}
\label{sec:alg:nmf}
Assuming we have a matrix $X \in \mathbb{R}^{D \times N}$. NMF allows us to factorize $X$ into two non-negative matrices $H \in \mathbb{R}^{D \times K}$ and $W \in \mathbb{R}^{K \times N}$, where the product of $H*W$ approximates original matrix $X$.

Each column in $X$ is a feature vector of the image with $D$ elements. Thus, $X$ represents the whole dataset with $N$ images. Conventionally, we name $H$ as basis matrix such that each column forms a basis vector of the semantic feature space, and $W$ as coefficient matrix. Hence, an image is further represented as the additive combination of weighted basis vectors in semantic space. $l_2$ norm distance (Equation~\ref{eq.l2}) and Kullback-Leibler (KL) divergence (Equation~\ref{eq.kl}) are two commonly-used measures of the similarity between original matrix $X$ and the product of $H$ and $W$, 
where $l_2$ norm distance is:
\begin{equation} \label{eq.l2}
\begin{aligned}
O^{l_2}&=\sum_{d=1}^D \sum_{n=1}^N (X_{dn}-\sum_{k=1}^K H_{dk}W_{kn})^2 = ||X-HW||^2 \\
		&=Tr(X^{\top}X)-2Tr(X^{\top}HW)+Tr(W^{\top}H^{\top}HW) 
\end{aligned}
\end{equation}

and Kullback-Leibler (KL) divergence is:
\begin{equation} \label{eq.kl}
O^{KL}=\sum_{d=1}^D \sum_{n=1}^N (X_{dn}ln\frac{X_{dn}}{(HW)_{dn}} - X_{dn} + (HW)_{dn})
\end{equation}

Based on different similarity measures, we are able to solve the factorization problem by minimizing the errors between $X$ and $H*W$. 

\subsection{MCC algorithm}
\label{sec:alg:mcc}

The authors in \cite{wang2013non} propose a new method to quantify the NMF by maximum correntropy criteria (MCC) for cancer clustering. Correntropy measures the generalized similarity between two random variables, $x$ and $y$, as defined in Equation~\ref{eq:correntropy}, where $k_\sigma$ is the kernel function and $E[.]$ is the expectation. 
\begin{equation} 
\label{eq:correntropy}
\hat{V}_\sigma (x-y) = E\left[ k_\sigma (x - y) \right]
\end{equation}

More precisely, correntropy models the expected differences between two random variables after mapping through kernel function. Unlike $l_2$ distance and KL divergence, such a modelling method can handle the nonlinear cases properly. Without knowing the joint distribution of $X$ and $Y$, we can simply estimate the expectation by taking a  average (shown in  Equation~\ref{eq.correntropyEstimation}):
\begin{equation} \label{eq.correntropyEstimation}
\hat{V}_\sigma (x-y) = \frac{1}{D} \sum_{i=1}^D k_\sigma (x_i - y_i)
\end{equation}

Instead of using $l_2$ distance or KL divergence as defined Equation~\ref{eq.l2} and~\ref{eq.kl}, respectively, we try to find the basis matrix $H$ and coefficient matrix $W$, whose product $Y$ is a well approximation of $X$, by maximizing their correntropy on a feature-by-feature basis to allow for weighting each feature differently. For each feature, the kernel function can be calculated as:
\begin{equation}
k_\sigma \left(\sqrt{\sum_{n=1}^N (x_{dn} - \sum_{k=1}^K h_{dk} w_{kn})^2}\right)
\end{equation}

Hence, the correntropy maximization problem is expressed as:
\begin{equation} \label{eq.mccKernel}
\max \limits_{h_{dk}>0,w_{kn}>0}\sum_{d=1}^D k_\sigma \left(\sqrt{\sum_{n=1}^N (x_{dn} - \sum_{k=1}^K h_{dk} w_{kn})^2}\right)
\end{equation}

To simplify the calculations without losing generality, they choose the Gaussian kernel function as $k_\sigma(.)$:
\begin{equation} \label{eq.gaussianKernel}
k_\sigma(x-y)=exp\left(-\gamma||x-y||^2\right)
\end{equation}

After substituting Equation~\ref{eq.gaussianKernel} back into Equation~\ref{eq.mccKernel}, the basis and coefficient matrices can be derived by solving:
\begin{equation}
\max \limits_{h_{dk}>0,w_{kn}>0} \sum_{d=1}^D exp \left(-\gamma \sum_{n=1}^N (x_{dn} - \sum_{k=1}^K h_{dk} w_{kn})^2\right)
\end{equation}

Then the convex conjugate function $\varphi(.)$ and auxiliary variables $\rho=[\rho_1,...,\rho_D]^\intercal$ are introduced to solve the above optimization problem based on the Proposition 1. According to the theory of convex conjugate functions, the above optimization problem is equivalent to:
\begin{equation}
\begin{aligned}
\max \limits_{H,W,\rho}  \quad   O^{MCC}(H,W,\rho)& \\
s.t. \quad H \geq 0, W \geq 0 &\nonumber \\
O^{MCC}(H,W,\rho)=\sum_{d=1}^D (& \rho_d \sum_{n=1}^{N} (x_{dn}-\sum_{k=1}^{K} h_{dk}w_{kn})^2\\
                                                                 &-\varphi(\rho_d) )\\
\end{aligned}
\end{equation}

\textbf{Proposition 1.} There exists a convex conjugate function of $g(z, \sigma)$ s.t.
\begin{equation}
\label{eq:ccf}
g(z, \sigma) = \sup_{\varrho \in \mathcal{R}^-} \Big( \varrho\frac{||z||^2}{\sigma^2} - \varphi(\varrho) \Big)
 \end{equation} 
and for a fixed $z$, the supremum is reached at $\varrho=-g(z,\sigma)$.
\subsection{GRNMF algorithm}
\label{sec:alg:grnmf}
The authors in \cite{cai2011graph} approach the nonlinear case based on the local invariance assumption \cite{belkin2001laplacian,niyogi2004locality} that if two images belong to the same topic, then they should be close to each other in both original (denoted as $x_n$ and $x_m$) and mapped feature spaces (denoted as $w_n$ and $w_m$). To model this intrinsic geometric information, manifold learning theory \cite{belkin2001laplacian} has proved to be effective by constructing a nearest neighbor graph on the data points. We can think the distribution of images as a graph, where each image is a node. If one $image_n$ is one of $k$ nearest neighbors of $image_m$ and vice versa, then we can build an edge between these two images with a weight. Hence, we can derive an affinity matrix $A$ where $A_{nm}$ indicates the weight between image $n$ and $m$. Then, the graph regularization is represented by Equation~\ref{eq:gr}.
\begin{equation}
\label{eq:gr}
\begin{aligned}
O^{GR}(W) &= \frac{1}{2}\sum_{n,m=1}^{N}||w_n-w_m||^2A_{nm} \\
				  &=Tr(WUW^{\intercal})-Tr(WAW^{\intercal}) \\
				  &=Tr(WLW^{\intercal})
\end{aligned}
\end{equation}
where $U$ is a diagonal matrix and $U_{nn}=\sum_{n=1}^{N}A_{nm}$, and $L=U-A$ is the graph Lapacian.

\subsection{MCCGR NMF algorithm}
\label{sec:alg:mccgr}

In this paper, we propose a new method that combines MCC algorithm and GRNMF. MCC algorithm shows good decomposition performance directly on the low-level features while the graph regularization can assist in this factorization process by considering the geometrical information of data. For convenience, we name our proposed method as MCCGR in this paper. 

The MCCGR algorithm tries to maximize the problem as defined in Equation~\ref{eq:mccgr}, where $\alpha$ is the weight of graph regularizer.
\begin{equation}
\label{eq:mccgr}
\begin{aligned}
\max \limits_{H,W,\rho}  \quad  & O^{MCCGR}(W) =  O^{MCC} - \alpha O^{GR}\\
s.t. \quad &H \geq 0, W \geq 0, \alpha \ge 0\\
\end{aligned}
\end{equation}

The optimization problem can be solved by Expectation-Maximization-like method. Starting from initial value of $H$ and $W$, $\rho$ is computed in expectation step (E-step). Conditional on the $\rho$ value, we can update the $H$ and $W$ values in maximization step (M-step). The process is called one iteration. This iterative process stops until it converges.

\textbf{Optimize $\rho$}: Starting from the estimated $H$ and $W$ from last M-step (or random values in the 1st iteration), $\rho$ of the \textit{t-th} iteration is computed as:
\begin{equation}
\label{eq:updateRho}
	\begin{aligned}
		&\rho_d^t=-g\left(\sqrt{\sum_{n=1}^N\left(x_{dn}-\sum_{k=1}^K h_{dk}^t w_{kn}^t\right)}, \sigma^t \right)& \\
	\end{aligned}	
\end{equation}
where 
\begin{equation}
\sigma^t = \sqrt{\frac{\theta}{2D} \sum_{d=1}^D \sum_{n=1}^N \left(x_{dn} - \sum_{k=1}^K h_{dk}^t w_{kn}^t \right)^2} \nonumber
\end{equation}

\textbf{Optimize $W$, $H$}: Conditional on the new $\rho$ from last step, we compute the new basis and coefficient matrix, denoted as $H^{t+1}$ and $W^{t+1}$ respectively, by maximizing the object function:
\begin{equation} \label{eq:primalMStep}
	\begin{aligned}	
		&(H^{t+1},W^{t+1}) \\
		= &\argmax_{H,W} \; \sum_{d=1}^D \left( \rho_d^t \sum_{n=1}^N \left(x_{dn}-\sum_{k=1}^K h_{dk} w_{kn} \right) ^2 \right) \\
		   &- \frac{1}{2}\alpha\sum_{n,m=1}^{N}||w_n-w_m||^2A_{nm} \\
		= &\argmax_{H,W} \; Tr[(X-HW)^\intercal diag(\rho^t)(X-HW)] \\
              &-\alpha Tr(WLW^{\intercal})\\
	\end{aligned}	
\end{equation}
where $diag(.)$ is the operator that converts a vector into a diagnal matrix.

According to the dual objective function, we have:
\begin{equation}
\label{eq:dualObj}
	\begin{aligned}	
		O^{D}(H, W)= &Tr[(X-HW)^\intercal diag(-\rho^t)(X-HW)] \\
              		&+\alpha Tr(WLW^{\intercal})\\
				=& Tr[X^\intercal diag(-\rho^t) X]\\
				  &-2Tr[X^\intercal diag(-\rho^t) HW] \\
				&+Tr[W^\intercal H^\intercal diag(-\rho^t)HW]\\
				 &+ \alpha Tr(WLW^{\intercal}) \\
\end{aligned}	
\end{equation}

Thus, the dual problem of Equation~\ref{eq:primalMStep} can be converted into following optimization problem:
\begin{equation} \label{eq:dualMStep}
	\begin{aligned}	
		(H^{t+1},W^{t+1}) &= \argmin_{H,W} \; O^{D}(H, W)\\
		s.t. &\quad H \ge 0, W \ge 0.
	\end{aligned}	
\end{equation}

We apply the Lagrange method to solve the optimization problem defined in Equation~\ref{eq:dualMStep}. Let the elements of matrices $\Phi=[\phi_{dk}]$ and $\Psi=[\psi_{kn}]$ be the corresponding Lagrange multipliers for the non-negative conditions of $h_{dk} \geq 0$ and $w_{kn} \geq 0$. Then we can express the Lagrange optimization problem as:
\begin{align}
\mathcal{L} &= Tr[X^\intercal diag(-\rho^t) X]-2Tr[X^\intercal diag(-\rho^t) HW] \nonumber \\
&+Tr[W^\intercal H^\intercal diag(-\rho^t) HW]  + Tr(WLW^{\intercal}) \nonumber \\
&+ Tr[\Phi H^\intercal] + Tr[\Psi H^\intercal]
\end{align}

The partial derivatives of $\mathcal{L}$ w.r.t. $H$ and $W$ are shown in Equation~\ref{eq:partialL}:
\begin{equation}
\label{eq:partialL}
\begin{aligned}
\frac{\partial \mathcal{L}}{\partial H}=&-2diag(-\rho^t)XW^\intercal + 2diag(-\rho^t)HWW^\intercal+\Phi \\
\frac{\partial \mathcal{L}}{\partial W}=&-2H^\intercal diag(-\rho^t)X+2H^\intercal diag(-\rho^t)HW \\
                                                   & + 2\alpha WL +\Psi \\
\end{aligned}
\end{equation}

Based on Karush-Kuhn-Tucker optimal conditions ($\phi_{dk}h_{dk}=0$ and $\psi_{kn}w_{kn}=0$), we can have:
\begin{equation}
\label{eq:partialL}
\begin{aligned}
&(diag(-\rho^t)HWW^\intercal)_{dk}h_{dk}-(diag(-\rho^t)XW^\intercal)_{dk}h_{dk} =0\\
&-(H^\intercal diag(-\rho^t)X)_{kn}w_{kn}+(H^\intercal diag(-\rho^t)HW)_{kn}w_{kn}\\
&+ \alpha (WL)_{kn}w_{kn} = 0\\
\end{aligned}
\end{equation}

Hence, the basis matrix $H$ and coefficient matrix $W$ are updated as follows:
\begin{equation}
\label{eq:updateH}
\begin{aligned}
h_{dk}^{t+1} &\leftarrow h_{dk}^t \frac{(diag(-\rho^t)XW^{t^\intercal})_{dk}}{(diag(-\rho^t)H^tW^tW^{t^\intercal}_{dk})_{dk}} \\
\end{aligned}
\end{equation}

\begin{equation}
\label{eq:updateW}
\begin{aligned}
w_{kn}^{t+1} &\leftarrow w_{kn}^t \frac{(H^{{t+1}^\intercal}diag(-\rho^t)X + \alpha WA)_{kn}}{(H^{{t+1}^\intercal}diag(-\rho^t)H^{t+1}W^t + \alpha WU)_{kn}}\\
\end{aligned}
\end{equation}

The overall iterative solution is summarized in the Algorithm~\ref{alg:update}.

\begin{algorithm}
\caption{MCCGR NMF Algorithm.}
\label{alg:update}
\begin{algorithmic}[l]
\State \textbf{INPUT}: $X$, Max\_iteration 
\State Initialize $H$ and $W$
\For {$t = 1 \colon$ Max\_iteration}
\State Update the auxiliary variables $\rho^{\intercal}$ from Equation~\ref{eq:updateRho}
\State Update $H^{t+1}$ from Equation~\ref{eq:updateH}
\State Update $W^{t+1}$ from Equation~\ref{eq:updateW}
\EndFor
\State \textbf{RETURN} $H^{t+1}$, $W^{t+1}$
\end{algorithmic}
\end{algorithm}

\subsection{Proof of convergence}
We now prove that the objective function $O^{D}$ in Equation~\ref{eq:dualObj} is non-increasing under the update rules defined in Equation~\ref{eq:updateH} and~\ref{eq:updateW}.

Firstly, we introduce the auxiliary function.

\textbf{Definition 1.} $G(w,w')$ is an auxiliary function for $F(w)$ if the following conditions hold:
\begin{equation}
G(w, w') \ge F(w), \quad G(w, w) = F(w)
\end{equation}

Next, we introduce a lemma that involves auxiliary function.

\textbf{Lemma 1.} If $G$ is an auxiliary function of $F$, then $F$ is non-increasing under the following update rule from iteration $t$ to $t+1$:
\begin{equation}
\label{eq:condition1}
w^{t+1} = \argmin_{w} G(w, w^{t})
\end{equation}

The proof of the Lemma 1 can be found in \cite{seung2001algorithms}. Thus, if we can find an auxiliary function for the objective function $O^{D}$ (Equation~\ref{eq:dualObj}) and an update rule that satisfies Lemma 1, we can show that $O^{D}$ is non-increasing.

We use $w_{kn}$ to represent any element in $W$, and $F_{kn}$ to represent the part of the $O^{D}$ that is only related to $w_{kn}$. We can derive following two equations:
\begin{equation}
\label{eq:first}
\begin{aligned}
F^{\prime}_{kn} = \left( \frac{\partial O^{D}}{\partial W}\right)_{kn} = &\Big( -2H^{\intercal} diag(-\rho^{\intercal})X \\ 
																	 &+ 2H^{\intercal}diag(-\rho^{\intercal})HW + 2\alpha WL\Big)_{kn} \\
\end{aligned}
\end{equation}
\begin{equation}
\label{eq:second}
F^{\prime\prime}_{kn} = \left( \frac{\partial^2 O^{D}}{\partial W^2}\right)_{kn} = \Big( 2H^{\intercal}diag(-\rho^{\intercal})H\Big)_{kk} + 2\alpha L_{nn}
\end{equation}

\textbf{Lemma 2.} The following function is an auxiliary function of $F_{kn}$ which is only relevant to variable $w_{kn}$:
\begin{equation}
\label{eq:g}
\begin{aligned}
G(w, &w^{t}_{kn})  =F^{t}_{kn}(w^{t}_{kn})  + F^{\prime}_{kn}(w^{t}_{kn})(w-w^{t}_{kn}) \\
					    &+\frac{\big(H^{\intercal}diag(-\rho^{\intercal})HW+2\alpha UW\big)_{kn}}{w^{t}_{kn}} (w-w^{t}_{kn})^2\\
\end{aligned}
\end{equation}

\textbf{Proof:} Obviously, $G(w, w) = F_{kn}(w)$. Now we show how to prove $G(w, w^{t}_{kn}) \ge F_{kn}(w)$. First, we expand the Taylor series of $F_{kn}(w)$ (as shown in Equation~\ref{eq:taloyF}).
\begin{equation}
\label{eq:taloyF}
\begin{aligned}
F_{kn}&(w) = F_{kn}(w^{t}_{kn}) + F^{\prime}_{kn}(w^{t}_{kn})(w-w^{t}_{kn}) \\
			&+ \frac{1}{2}F^{\prime\prime}_{kn}(w^{t}_{kn})(w-w^{t}_{kn})^2\\
			=&F_{kn}(w^{t}_{kn}) + F^{\prime}_{kn}(w^{t}_{kn})(w-w^{t}_{kn}) \\
			&+ \big((H^{\intercal}diag(-\rho^{\intercal})H)_{kk} + \alpha L_{nn}\big)(w-w^{t}_{kn})^2\\
\end{aligned}
\end{equation}

To show $G(w, w^{t}_{kn}) \ge F_{kn}(w)$, we need to prove that the following relation holds:
\begin{equation}
\label{eq:assumption}
\begin{aligned}
\frac{(H^{\intercal}diag(-\rho^{\intercal})HW + 2\alpha UW)_{kn}}{w^{t}_{kn}} \ge &(H^{\intercal}diag(-\rho^{\intercal})H)_{kk} \\
																						 &+ (\alpha L)_{nn}
\end{aligned}
\end{equation}

We have
\begin{equation}
\begin{aligned}
(H^{\intercal}diag(-\rho^{\intercal})HW)_{kn} &= \sum_{l=1}^{k}(H^{\intercal}diag(-\rho^{\intercal})H)_{kl}w^{t}_{ln}\\
												&\ge (H^{\intercal}diag(-\rho^{\intercal})H)_{kk}w^{t}_{kn}
\end{aligned}
\end{equation}

and
\begin{equation}
\begin{aligned}
\alpha (UW)_{kn} = \alpha \sum_{j=1}^{D}(U_{nj}w^{t}_{jk} &\ge \alpha U_{nn}w^{t}_{kn}\\
																    &\ge \alpha (U-A)_{nn}w^{t}_{kn}\\
																    &= \alpha L_{nn}w^{t}_{kn}\\
\end{aligned}
\end{equation}

Hence, inequality~\ref{eq:assumption} holds and $G(w, w^{t}_{kn}) \ge F_{kn}(w)$.

If we replace $G(w, w^{t}_{kn})$ in \ref{eq:condition1} by the results of~\ref{eq:g}, we have following update rule:
\begin{equation}
\begin{aligned}
w^{t+1}_{kn} &= w^{t}_{kn} - w^{t}_{kn} \frac{F^{\prime}_{kn}(w^{t}_{kn})}{2(H^{\intercal}diag(-\rho^{\intercal})HW + 2\alpha UW^{t})_{kn}}\\
				  &= w^{t}_{kn} \frac{2(H^{\intercal}diag(-\rho^{\intercal})X + 2\alpha AW)_{kn}}{(H^{\intercal}diag(-\rho^{\intercal})HW^{t} + \alpha UW^{t})_{kn}}\\
\end{aligned}
\end{equation}

Thus, $G(w, w^{t}_{kn})$ (shown in Equation~\ref{eq:g}) is an auxiliary function for $F_{kn}(w)$, and $F_{kn}(w)$ is non-increasing under this update rule.

In a similar vein, we can show that $O^D$ is non-increasing under the update rule in Equation~\ref{eq:updateH}.

\section{Experiment Settings}
\label{sec:expSettings}
We use two datasets for testing: Caltech101 \cite{fei2007learning} and Caltech256 \cite{griffin2007caltech}. We follow the way in \cite{gehler2009feature} to extract features. First, SIFT algorithm \cite{lowe1999object} is used to extract the local invariant features of each image in grayscale. Then all SIFT descriptors of both datasets are gathered together to form a dictionary, which is used for vector quantization for all images. Two types of dictionaries are used with different numbers of codewords: 300 and 1000. Thus, each image is represented as an either 300- or 1000-dimension feature vector.

We randomly select $K$ categories and use all images from these $K$ categories to form the testing matrix $X$. Intuitively, the value of $K$ controls the decomposition of the matrix. Fundamentally, it determines the number of basis vectors that span the latent semantic space. To fully investigate the efficiency of the proposed algorithm, the number of cluster ranges from 1 to 10 for both datasets. We repeat this process 50 times to reduce the potential effect of random errors in experiment results, since the performance of NMF algorithm is affected by the initial values of the iterative process. The $H$ and $W$ are initialized by random values. The output of each algorithm is derived from taking the average of 50 runs. 

To evaluate the decomposition performance, we directly apply \textit{k}-means clustering method to sub-group $W$ into $K$ clusters. Essentially, each column of $W$ with $K$ elements are the new feature vector of one image in the latent semantic spaces after matrix decomposition. \textit{K}-means will assign each image with a label. We compare the label from \textit{k}-means to the original ground-truth label to evaluate the clustering results. The performance is measured by accuracy (shown in Equation ~\ref{eq:accu}). To find the correspondence between the class from ground-truth data and the label by \textit{k}-means,  we use Kuhn-Munkres algorithm \cite{kuhn1955hungarian}.
\begin{equation}
\label{eq:accu}
Accuracy = \frac{\sum\limits_{i=1}^{N} \delta(kmeans\_label_i, class_i)}{N}
\end{equation}
where $\delta(kmeans\_label_i, class_i)$ returns 1 if $kmeans\_label_i = class_i$; otherwise 0. 

Besides accuracy, we also use normalized mutual information (NMI) as a measure. Mutual information is to measure how close two clusters are. Given a randomly selected image and two clusters of images, $C_1$ and $C_2$, we denote $p(C_1)$ and $p(C_2)$ as the probability that this image belongs to $C_1$ and $C_2$, respectively, and $p(C_1, C_2)$ as the joint probability that this image is in both clusters at the same time. Then, the mutual information $MI(C_1, C_2)$ is defined as follows:
\begin{equation}
MI(C_1, C_2)=\sum_{c_i \in C_1, c_j \in C_2} p(c_i, c_j) log \frac{p(c_i, c_j)}{p(c_i)p(c_j)}
\end{equation}

Similar to \cite{cai2011graph}, we adopt the normalized mutual information (defined in Equation.~\ref{eq:nmi}) by considering the entropy. 
\begin{equation}
\label{eq:nmi}
NMI(C_1, C_2)=\frac{MI(C_1, C_2)}{max(H(C_1), H(C_2))}
\end{equation}

where $H(C_1)$ is the entropy of $C_1$.

\section{Results and Discussions}
\label{sec:results}

\subsection{Parameter selection}
\label{subsec:alpha}

\begin{figure}[ht!] \centering
\subfigure[Caltech101, 300 codewords] { \label{subfig:101_k300}
\includegraphics[width=0.46\columnwidth]{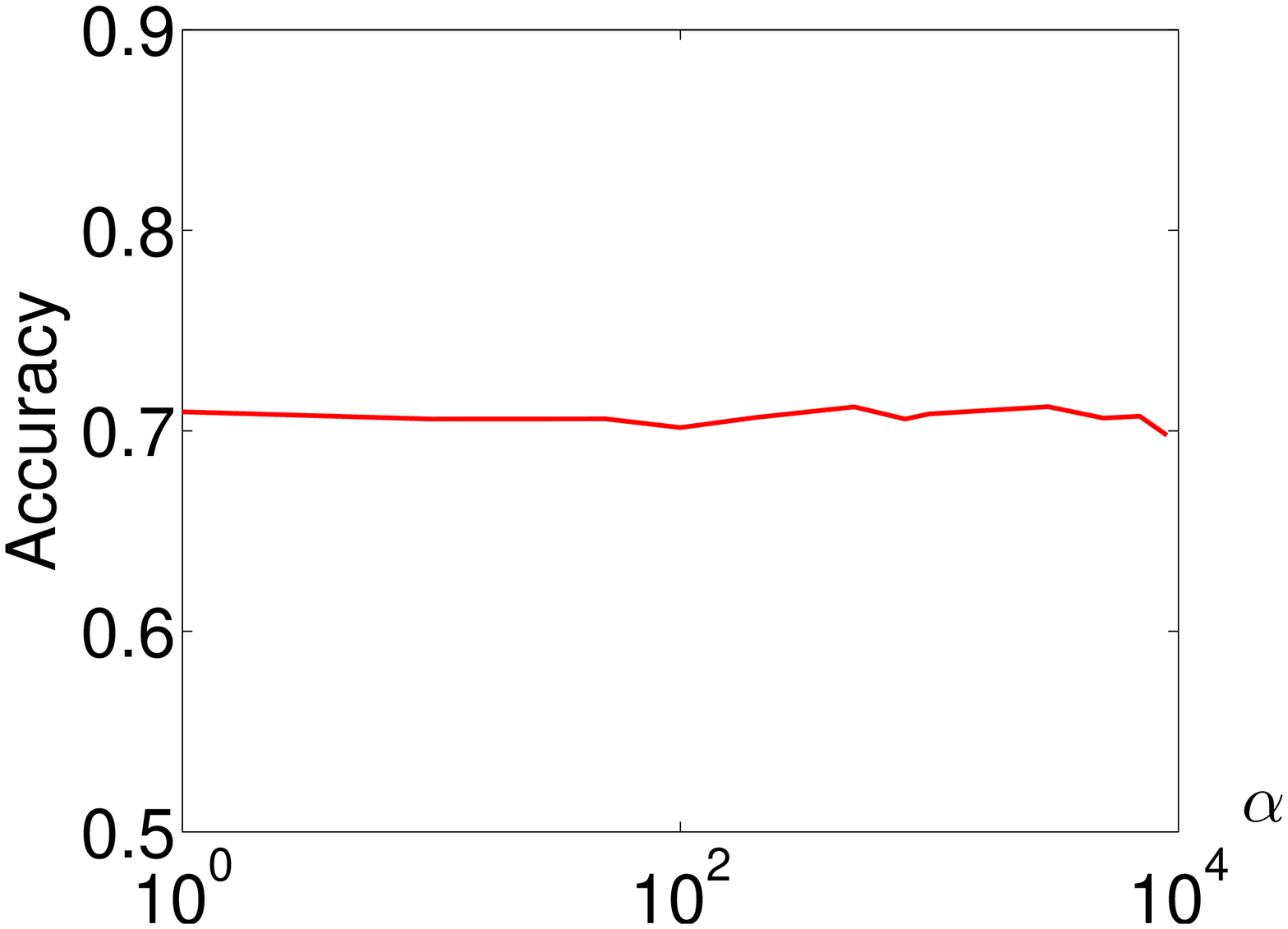}
}
\subfigure[Caltech256, 300 codewords] { \label{subfig:256_k300}
\includegraphics[width=0.46\columnwidth]{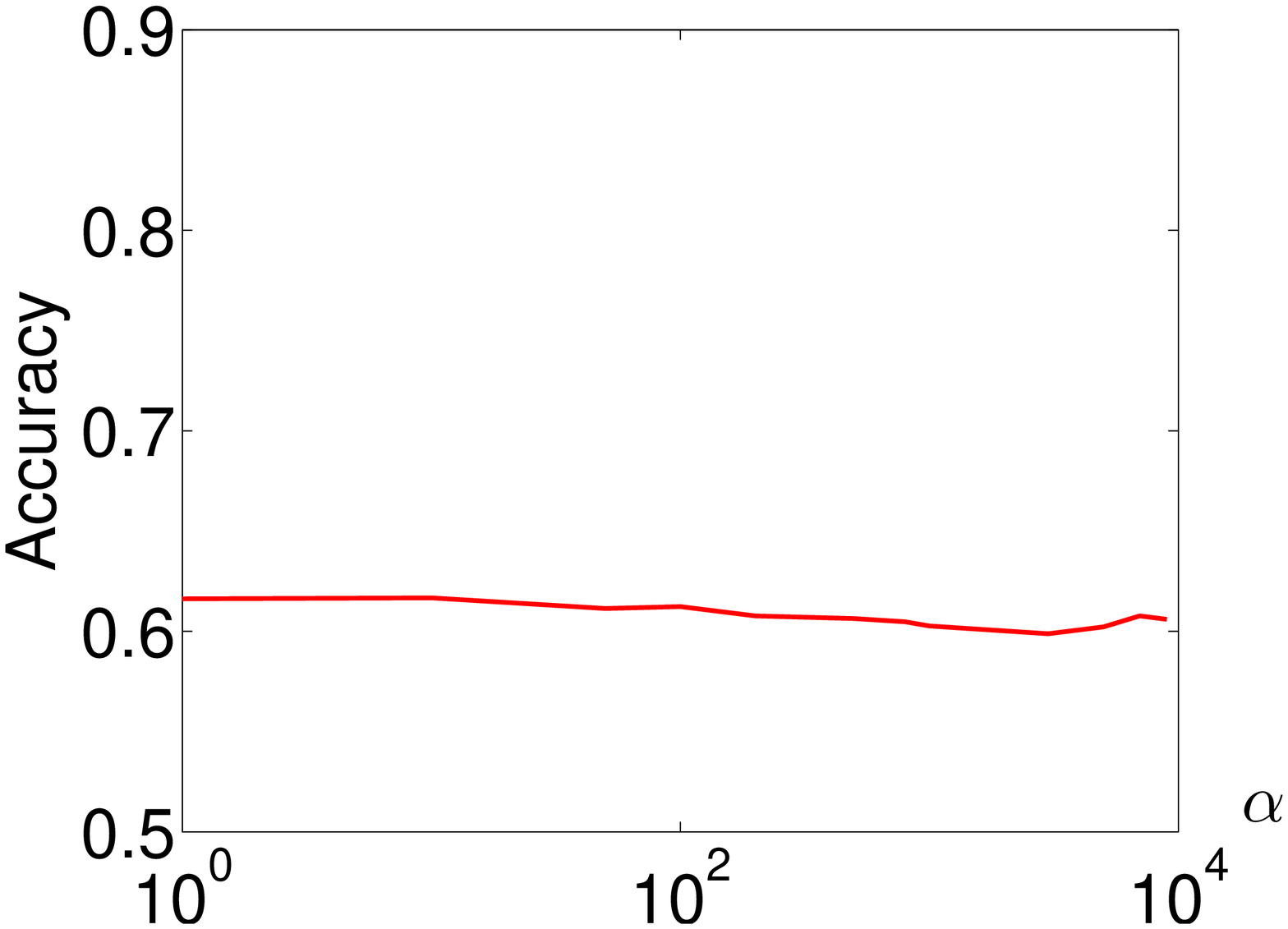}
}
\subfigure[Caltech101, 1000 codewords] { \label{subfig:101_k1000}
\includegraphics[width=0.46\columnwidth]{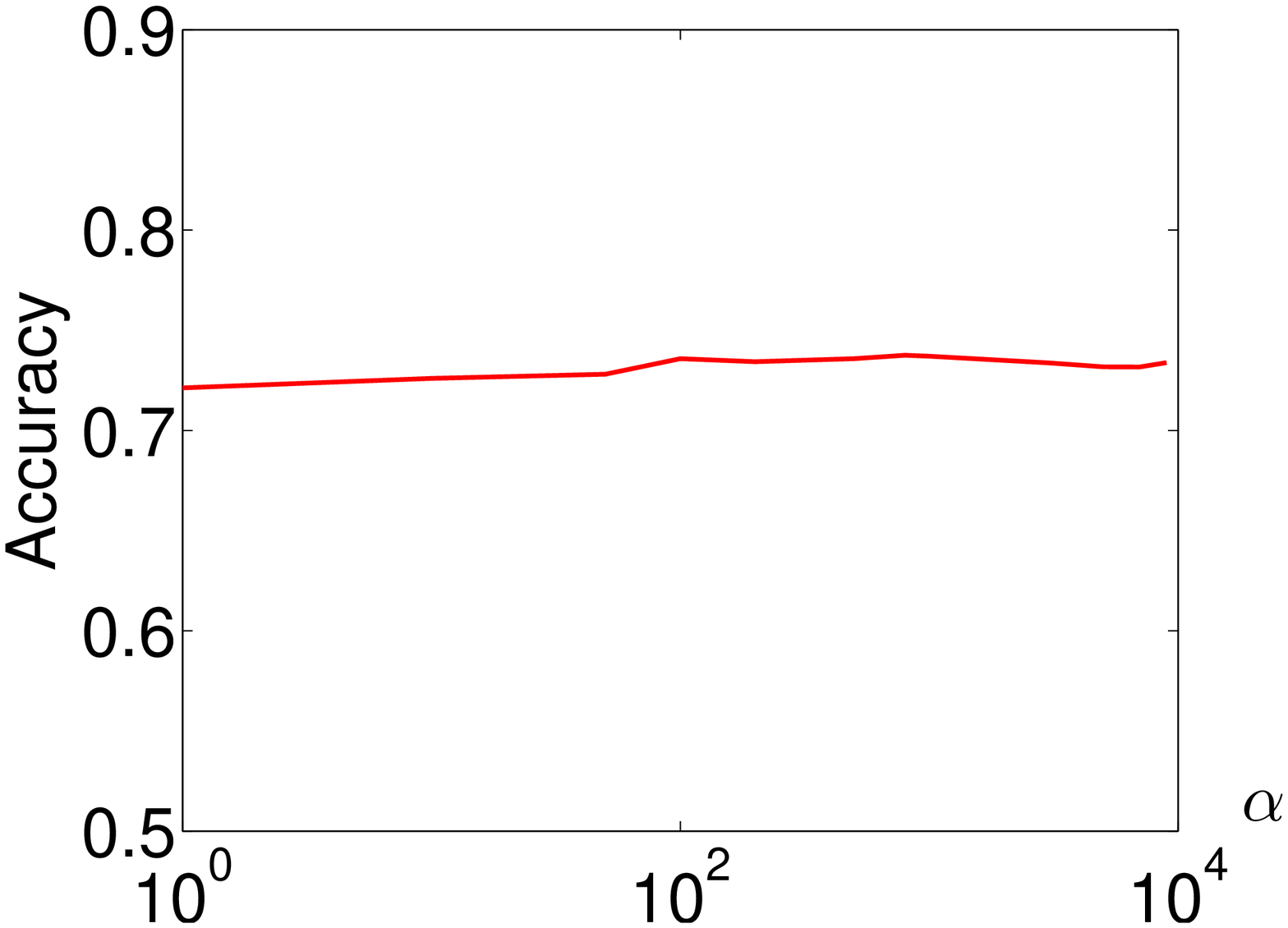}
}
\subfigure[Caltech256, 1000 codewords] { \label{subfig:256_k1000}
\includegraphics[width=0.46\columnwidth]{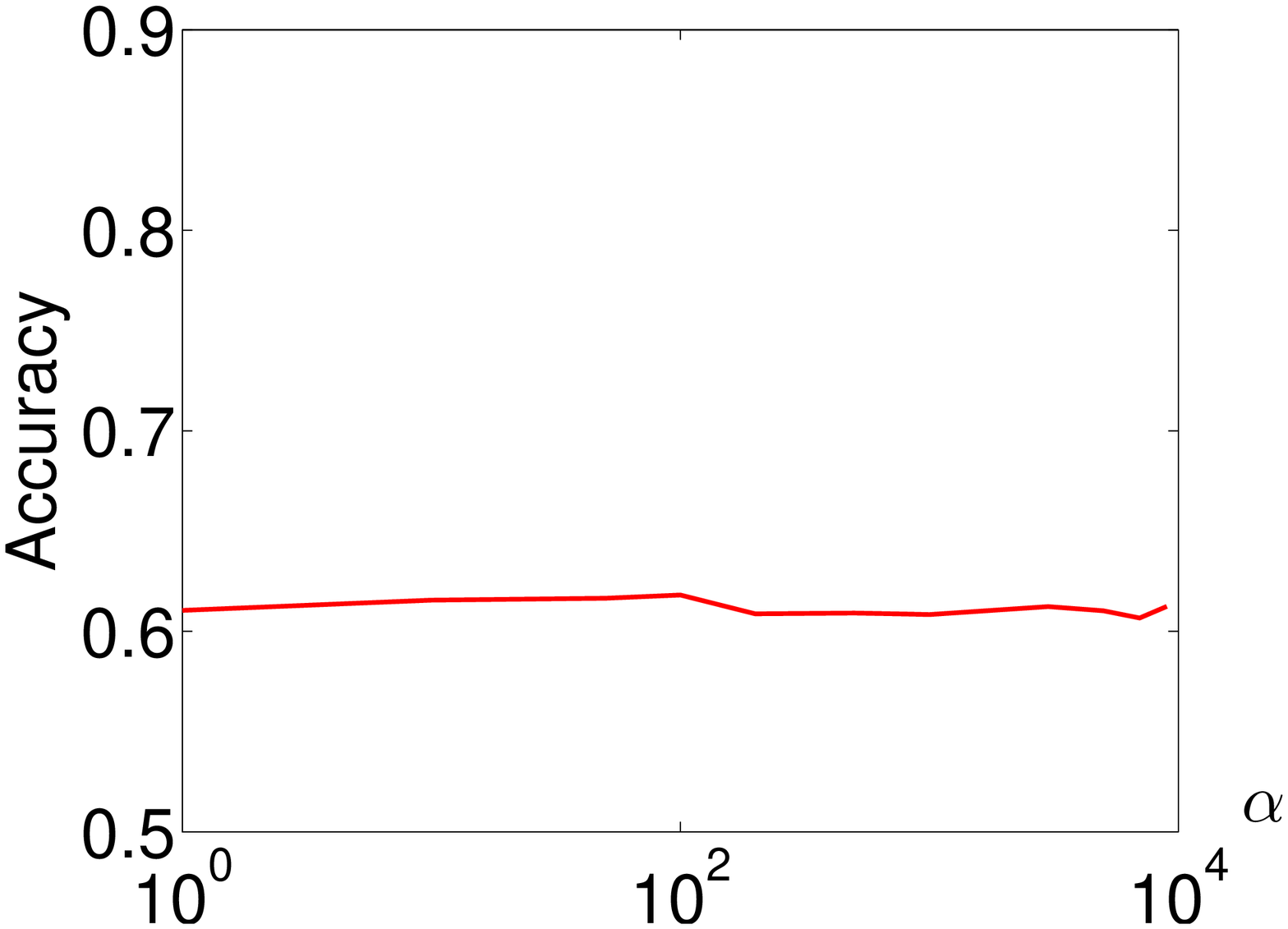}
}
\caption{The effects of $\alpha$ of MCCGR algorithm on the accuracy.}
\label{fig:alpha}
\end{figure}

One important parameter we need to highlight is the weight of graph regularization ($\alpha$). $\alpha$ indicates the relative importance of graph regularization in our decomposition process. Thus it's worthwhile to systematically investigate the effects of $\alpha$ on the accuracy.

We vary the values of $\alpha$ in the range of [1, 10000] and fix $K=2$ (i.e. two-class clustering). We randomly choose 2 classes of images for testing with different $\alpha$, and then repeat this process 50 times to reduce the potential effects of randomness in our experiment. The average of the 50 runs can be found in Figure~\ref{fig:alpha}. 

In this figure, $x$-axis in log-scale demonstrates the choice of $\alpha$, while $y$-axis indicates the corresponding accuracy. As we can see, the MCCGR algorithm is robust enough to the change of $\alpha$. The accuracy fluctuates in a small range as $\alpha$ increases from 1 to 10000. 

\subsection{Convergence study of MCCGR NMF}
\label{sec:result:proof}
\begin{figure}[h!] \centering
\subfigure[Caltech101, 300 codewords] { \label{subfig:101_k300}
\includegraphics[width=0.46\columnwidth]{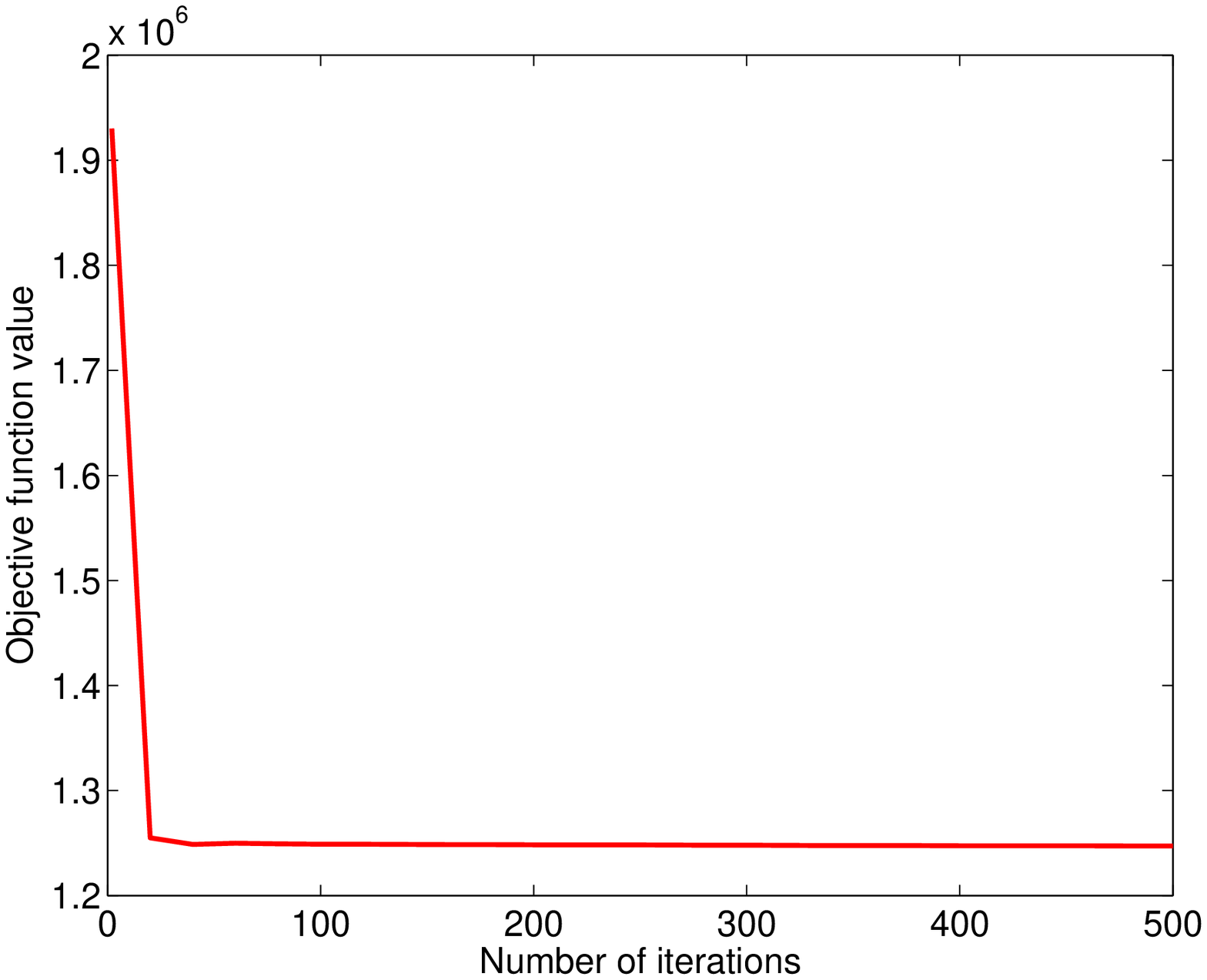}
}
\subfigure[Caltech256, 300 codewords] { \label{subfig:256_k300}
\includegraphics[width=0.46\columnwidth]{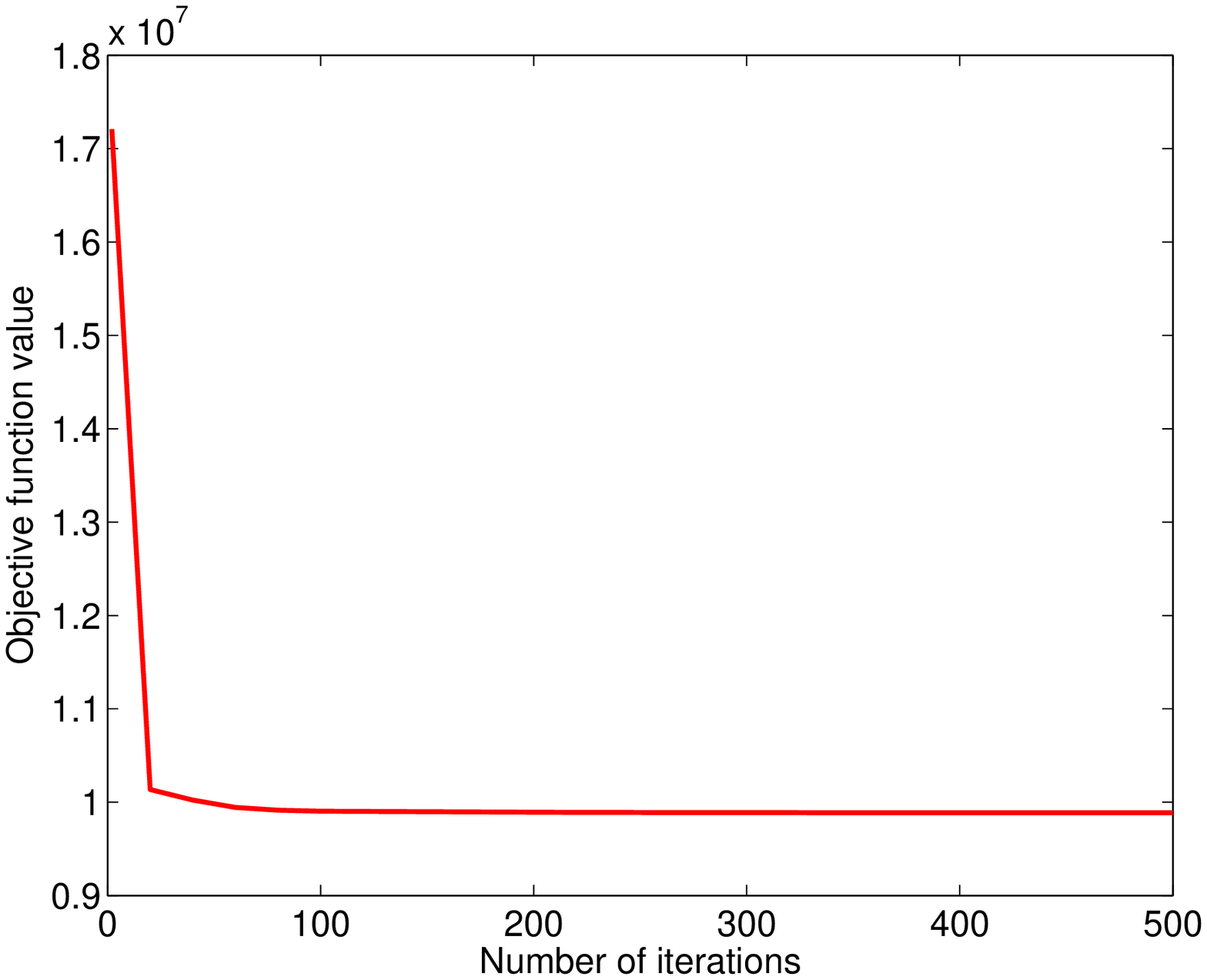}
}
\subfigure[Caltech101, 1000 codewords] { \label{subfig:101_k1000}
\includegraphics[width=0.46\columnwidth]{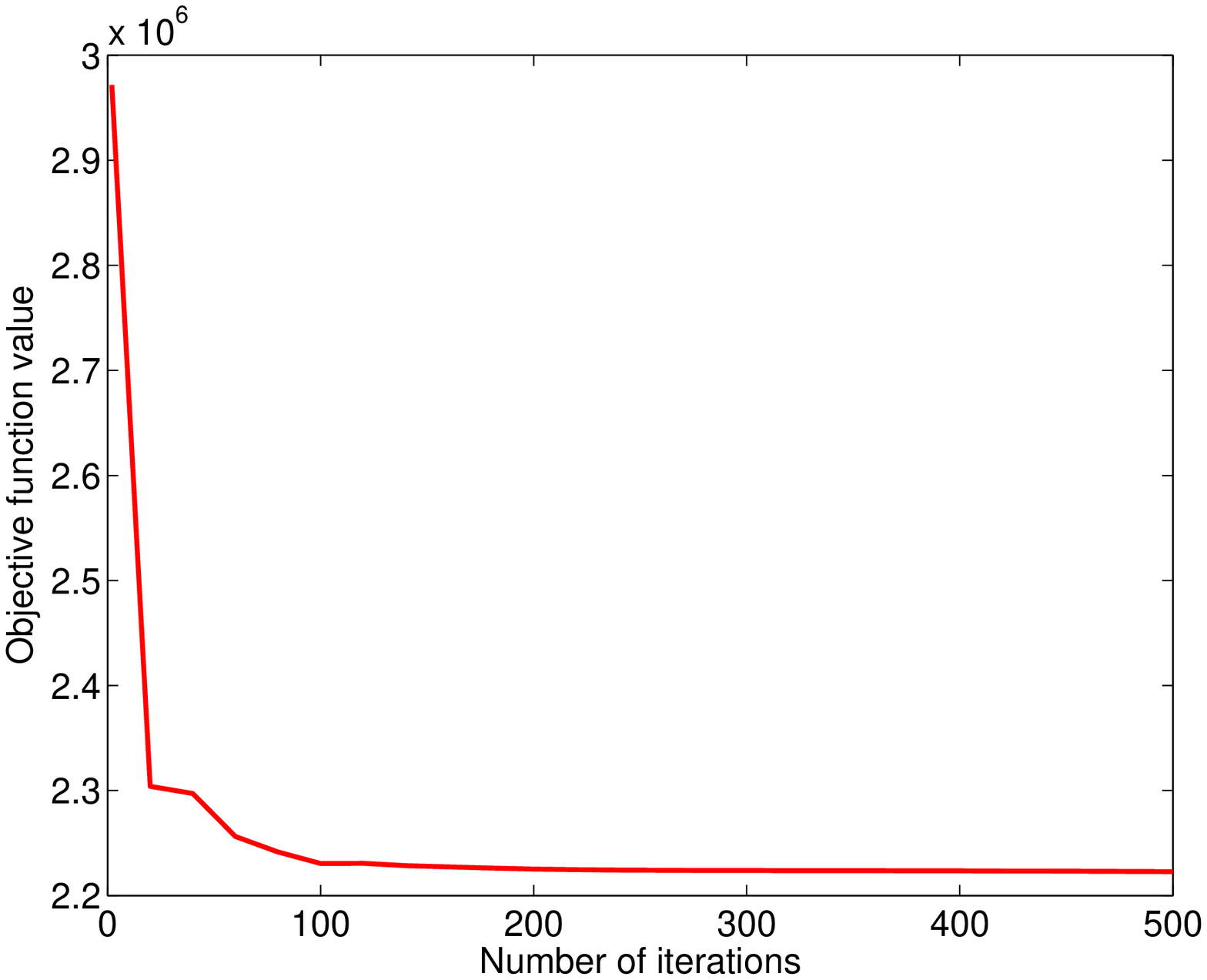}
}
\subfigure[Caltech256, 1000 codewords] { \label{subfig:256_k1000}
\includegraphics[width=0.46\columnwidth]{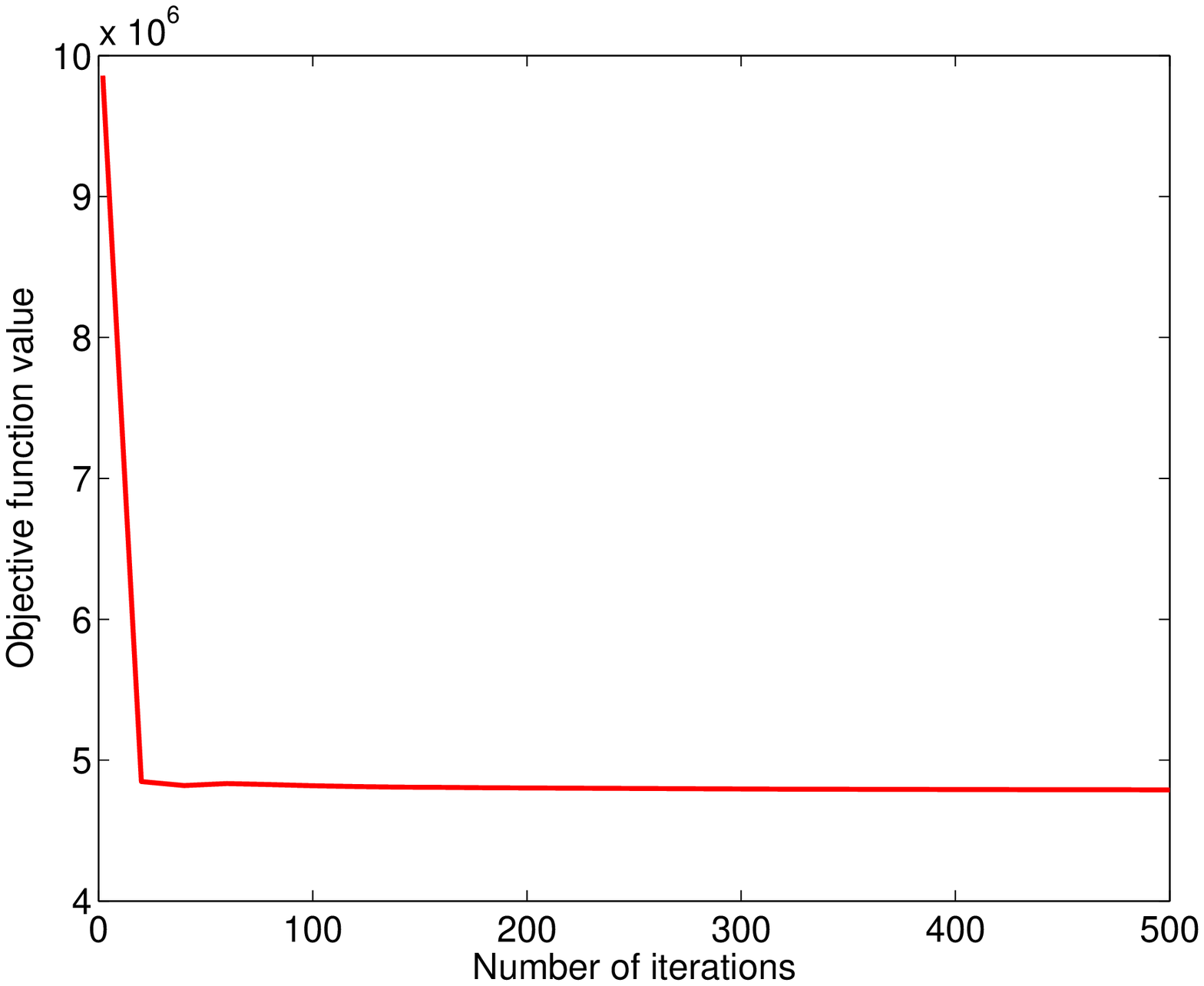}
}
\caption{Convergence curves of MCCGR algorithm on Caltech101 and Caltech256 datasets, with 300- and 1000- codewords options.}
\label{fig:converge}
\end{figure}

We also plot the convergence curves (Figure~\ref{fig:converge}) of our proposed MCCGR algorithm for both datasets with different codeword settings. The $y$-axis shows the objective function value while $x$-axis means the iteration number. As we can see from the figures, the proposed algorithm converges quickly within less than 100 iterations. In most cases, it needs less than 50 iterations.

\subsection{Comparison with other NMF algorithms}
\label{subsec:algs}
\begin{figure*}[ht!] \centering
\subfigure[Caltech101, 300 codewords] { \label{subfig:101_k300}
\includegraphics[width=0.68\columnwidth]{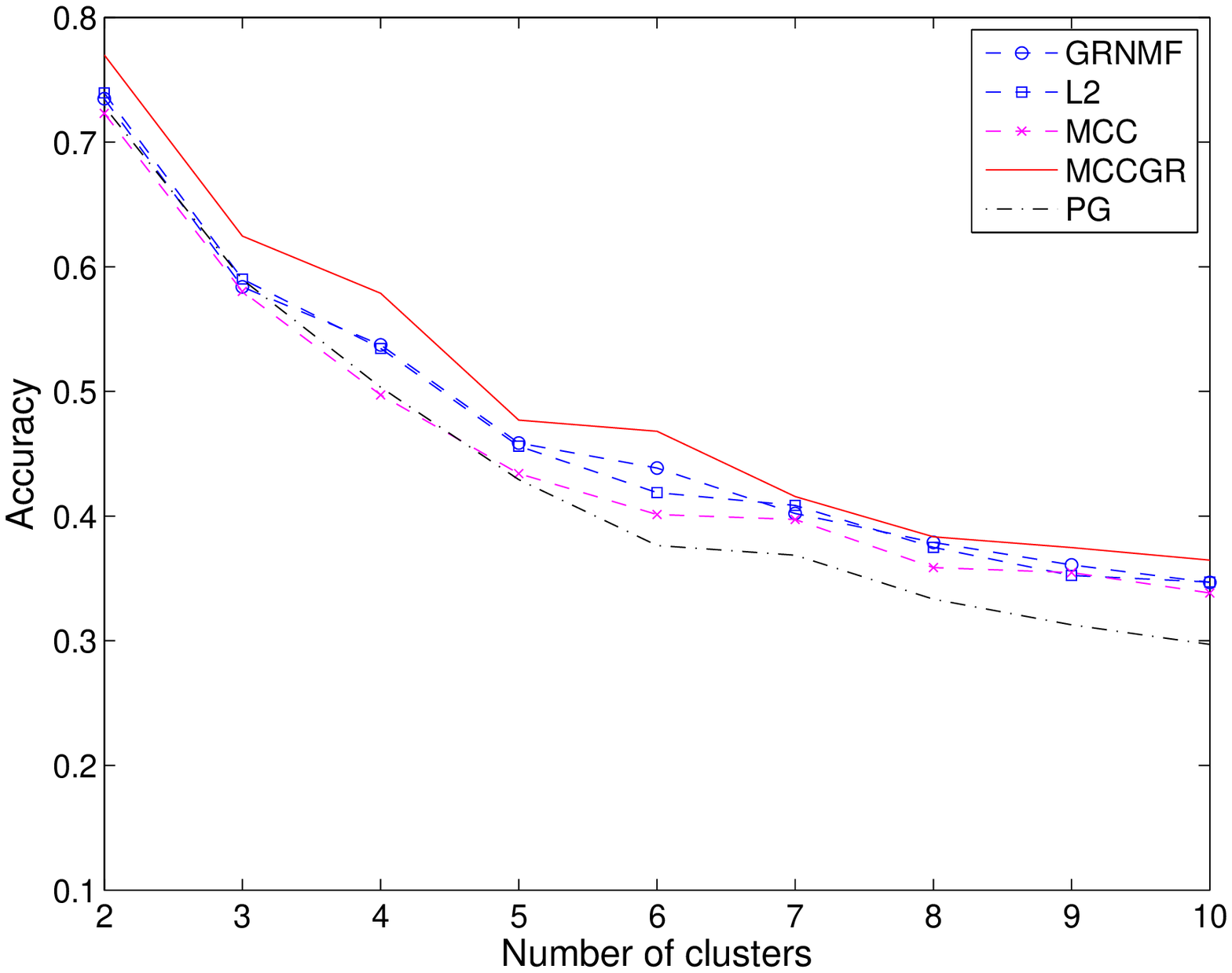}
}
\subfigure[Caltech256, 300 codewords] { \label{subfig:256_k300}
\includegraphics[width=0.68\columnwidth]{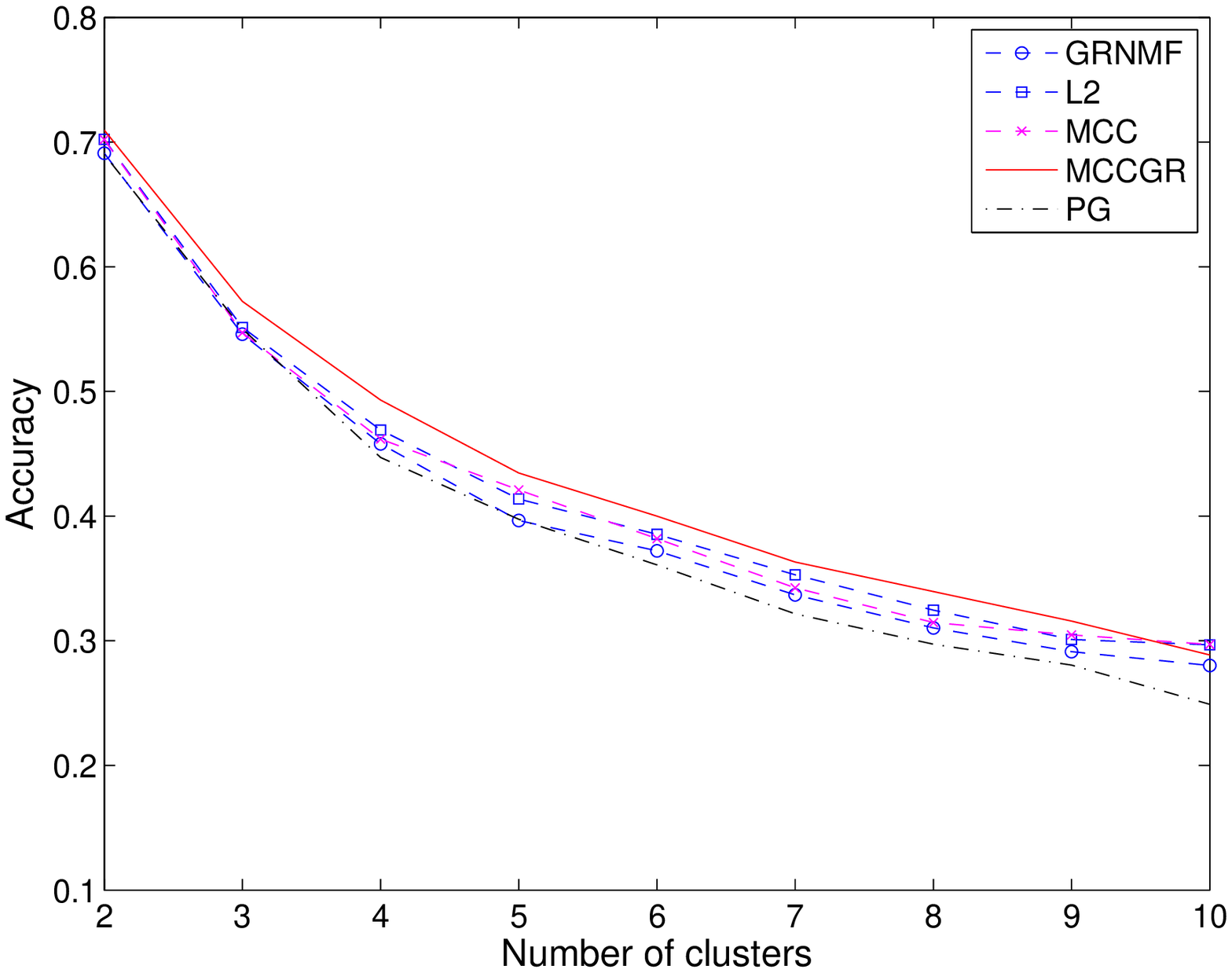}
}
\subfigure[Caltech101, 1000 codewords] { \label{subfig:101_k1000}
\includegraphics[width=0.68\columnwidth]{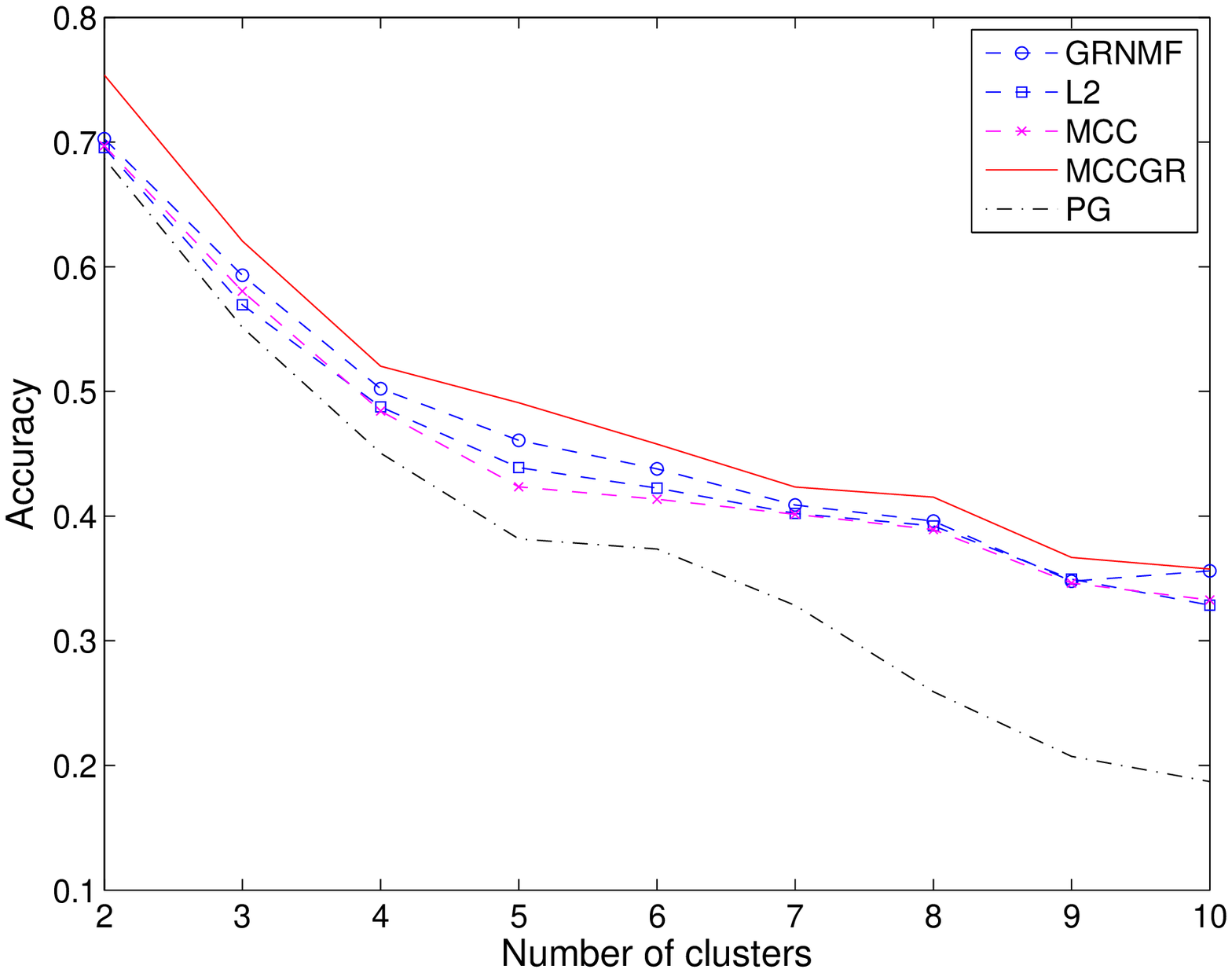}
}
\subfigure[Caltech256, 1000 codewords] { \label{subfig:256_k1000}
\includegraphics[width=0.68\columnwidth]{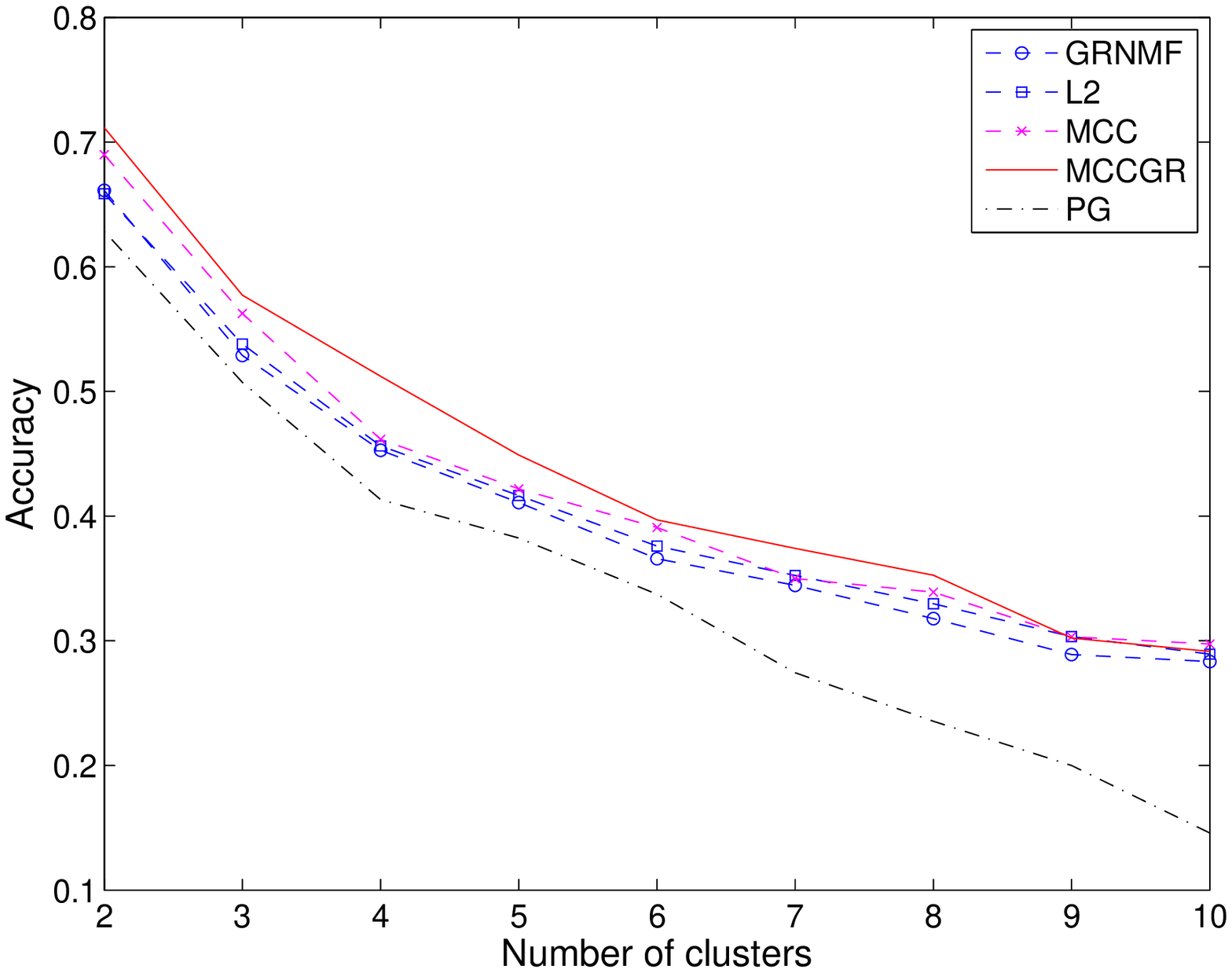}
}
\caption{Accuracies on Caltech101 and Caltech256 datasets, with 300- and 1000- codewords.}
\label{fig:accu}
\end{figure*}

\newcommand{\tabincell}[2]{\begin{tabular}{@{}#1@{}}#2\end{tabular}}
\begin{table*}[ht]
\centering
\caption{Clustering accuracies on Caltech101 and Caltech256 datasets, with 300 codewords in dictionary} 
\label{tab.results300}
\begin{tabular}{|c||c|c|c|c|c||c|c|c|c|c|}
\hline
\multirow{2}{*}{\tabincell{c}{K}} & \multicolumn{5}{|c||}{Caltech101} & \multicolumn{5}{|c|}{Caltech256}\\
\hhline{~----------}
 & GRNMF &  L2 & MCC & MCCGR & PG  & GRNMF & L2 & MCC & MCCGR & PG \\ \hline
2 & 0.735 & 0.739 & 0.723 & \bf{0.770} & 0.728   & 0.691 & 0.702 & 0.702 & \bf{0.710} & 0.690\\ \hline 
3 & 0.584 & 0.590 & 0.580 & \bf{0.625} & 0.590  & 0.546 & 0.551 & 0.547 & \bf{0.572} & 0.550 \\ \hline 
4 & 0.537 & 0.535 & 0.497 & \bf{0.579} & 0.503  & 0.458 & 0.469 & 0.462 & \bf{0.493} & 0.447  \\ \hline 
5 & 0.459 & 0.456 & 0.434 & \bf{0.477} & 0.429 & 0.396 & 0.414 & 0.421 & \bf{0.435} & 0.397  \\ \hline 
6& 0.439 & 0.419 & 0.401 & \bf{0.468} & 0.376 & 0.372 & 0.385 & 0.382 & \bf{0.400} & 0.361   \\ \hline 
7 & 0.402 & 0.409 & 0.397 & \bf{0.416} & 0.369 & 0.337 & 0.353 & 0.342 & \bf{0.363} & 0.322   \\ \hline 
8 & 0.379 & 0.375 & 0.359 & \bf{0.383} & 0.333   & 0.310 & 0.325 & 0.315 & \bf{0.339} & 0.297  \\ \hline 
9 & 0.361 & 0.352 & 0.355 & \bf{0.375} & 0.313  & 0.291 & 0.301 & 0.305 & \bf{0.316} & 0.280  \\ \hline 
10 & 0.339 & 0.325 & 0.324 & \bf{0.343} & 0.278  & 0.280 & \bf{0.297} & \bf{0.297} & 0.289 & 0.249  \\ \hline 
\end{tabular}
\end{table*}

\begin{table*}[ht]
\centering
\caption{Clustering accuracies on Caltech101 and Caltech256 datasets, with 1000 codewords in dictionary} \label{tab.results1000}
\begin{tabular}{|c||c|c|c|c|c||c|c|c|c|c|}
\hline
\multirow{2}{*}{\tabincell{c}{K}} & \multicolumn{5}{|c||}{Caltech101} & \multicolumn{5}{|c|}{Caltech256}\\
\hhline{~----------}
 & GRNMF &  L2 & MCC & MCCGR & PG  & GRNMF & L2 & MCC & MCCGR & PG \\ \hline
2 & 0.703 & 0.696 & 0.697 & \bf{0.754} & 0.687   & 0.661 & 0.659 & 0.690 & \bf{0.712} & 0.628 \\ \hline 
3 & 0.593 & 0.569 & 0.580 & \bf{0.621} & 0.551 & 0.529 & 0.538 & 0.563 & \bf{0.577} & 0.507  \\ \hline 
4 & 0.502 & 0.488 & 0.484 &\bf{0.520} & 0.450 & 0.453 & 0.456 & 0.462 & \bf{0.512} & 0.413   \\ \hline 
5 & 0.461 & 0.439 & 0.423 & \bf{0.491} & 0.382 & 0.411 & 0.417 & 0.422 & \bf{0.449} & 0.382   \\ \hline 
6 & 0.438 & 0.423 & 0.414 & \bf{0.458} & 0.374  & 0.366 & 0.376 & 0.391 & \bf{0.397} & 0.337   \\ \hline 
7 & 0.409 & 0.402 & 0.402 & \bf{0.423} & 0.328 & 0.344 & 0.352 & 0.350 & \bf{0.374} & 0.274 \\ \hline 
8 & 0.396 & 0.392 & 0.389 & \bf{0.415} & 0.259  & 0.318 & 0.330 & 0.339 & \bf{0.353} & 0.235\\ \hline 
9 & 0.348 & 0.349 & 0.346 & \bf{0.367} & 0.207 & 0.289 & \bf{0.303} & \bf{0.303} & 0.302 & 0.200  \\ \hline 
10 & 0.356 & 0.329 & 0.333 & \bf{0.358} & 0.187 & 0.283 & 0.289 & \bf{0.297} & 0.291 & 0.146  \\ \hline 
\end{tabular}
\end{table*}

\begin{figure*}[ht!] \centering
\subfigure[Caltech101, 300 codewords] { \label{subfig:101_k300}
\includegraphics[width=0.68\columnwidth]{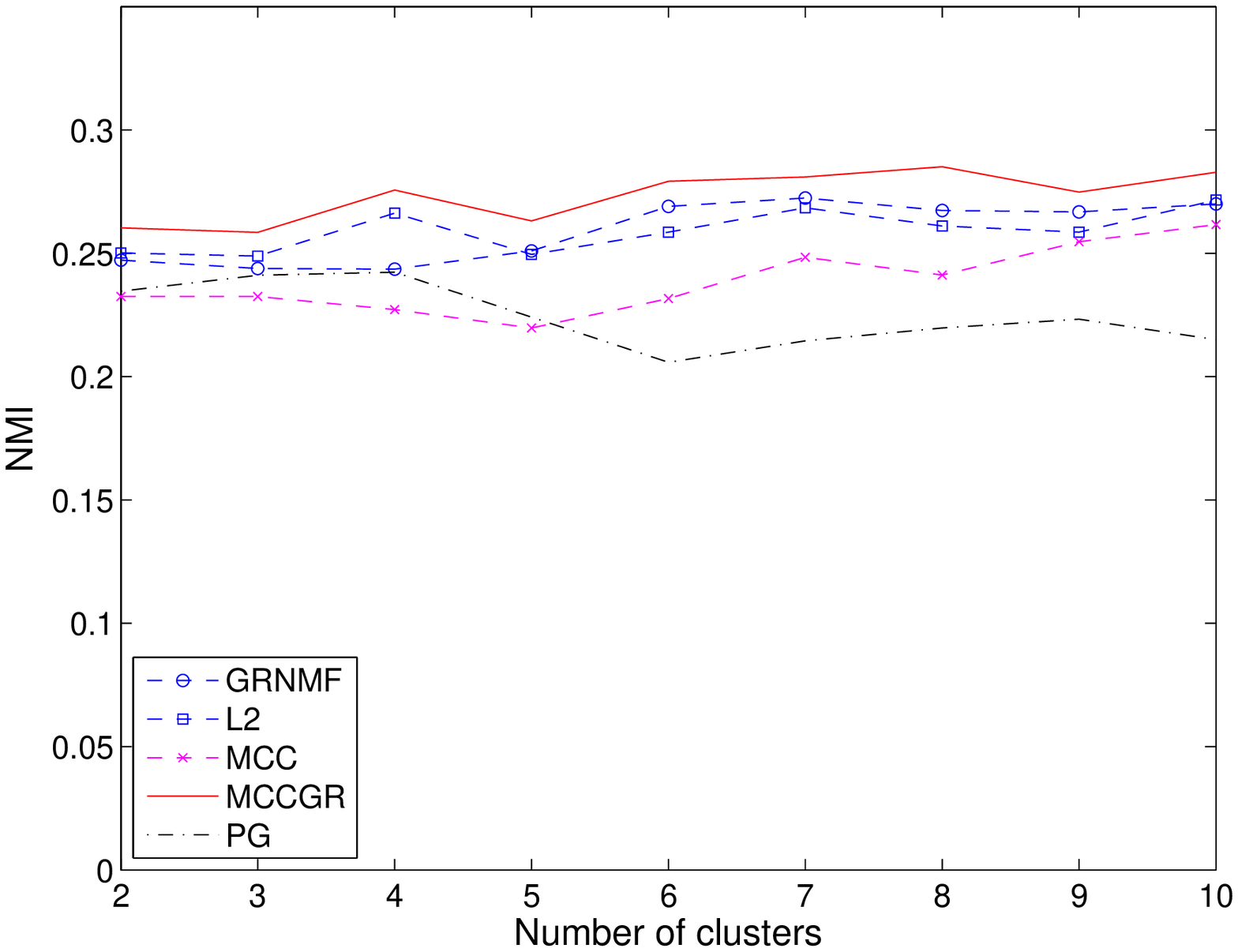}
}
\subfigure[Caltech256, 300 codewords] { \label{subfig:256_k300}
\includegraphics[width=0.68\columnwidth]{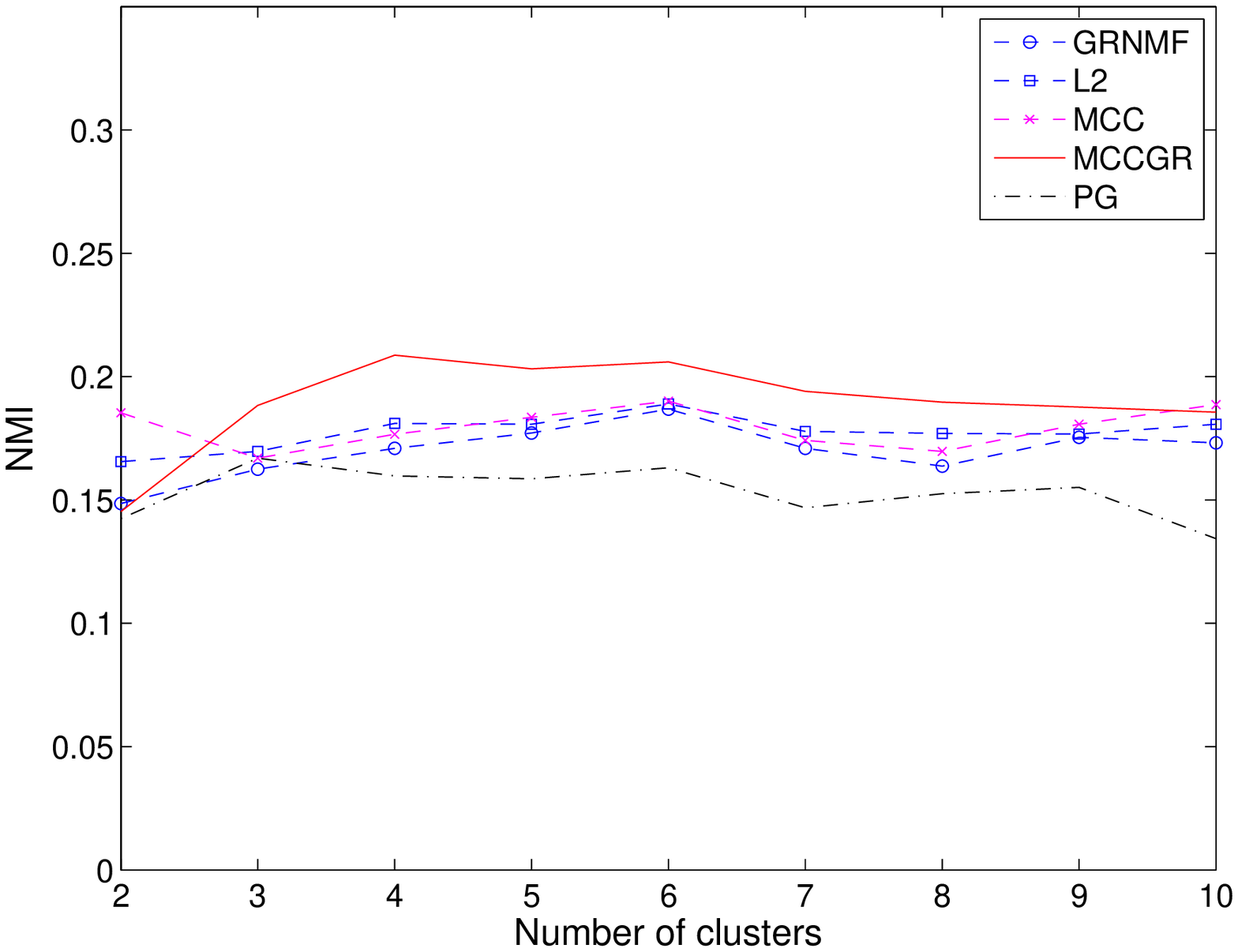}
}
\subfigure[Caltech101, 1000 codewords] { \label{subfig:101_k1000}
\includegraphics[width=0.68\columnwidth]{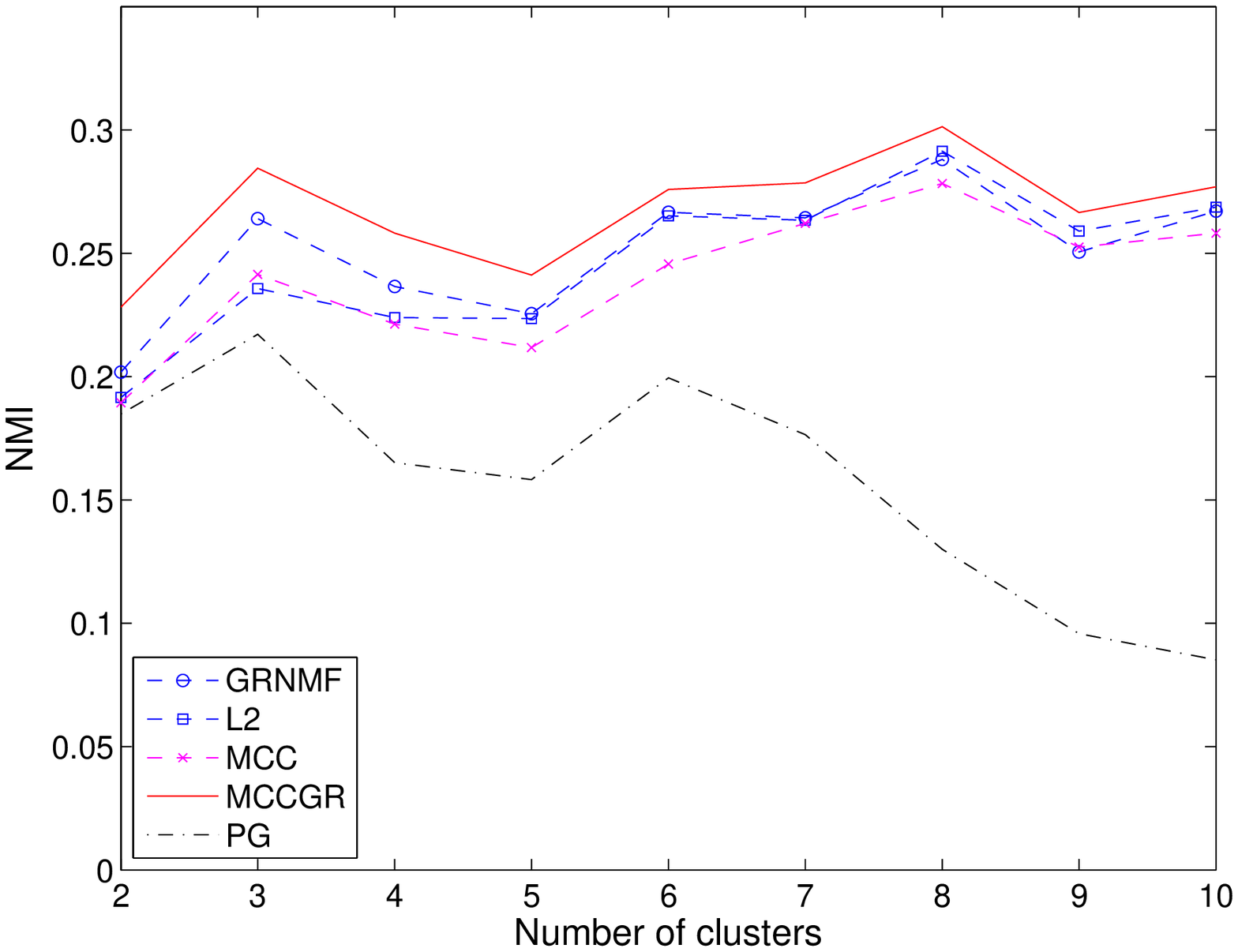}
}
\subfigure[Caltech256, 1000 codewords] { \label{subfig:256_k1000}
\includegraphics[width=0.68\columnwidth]{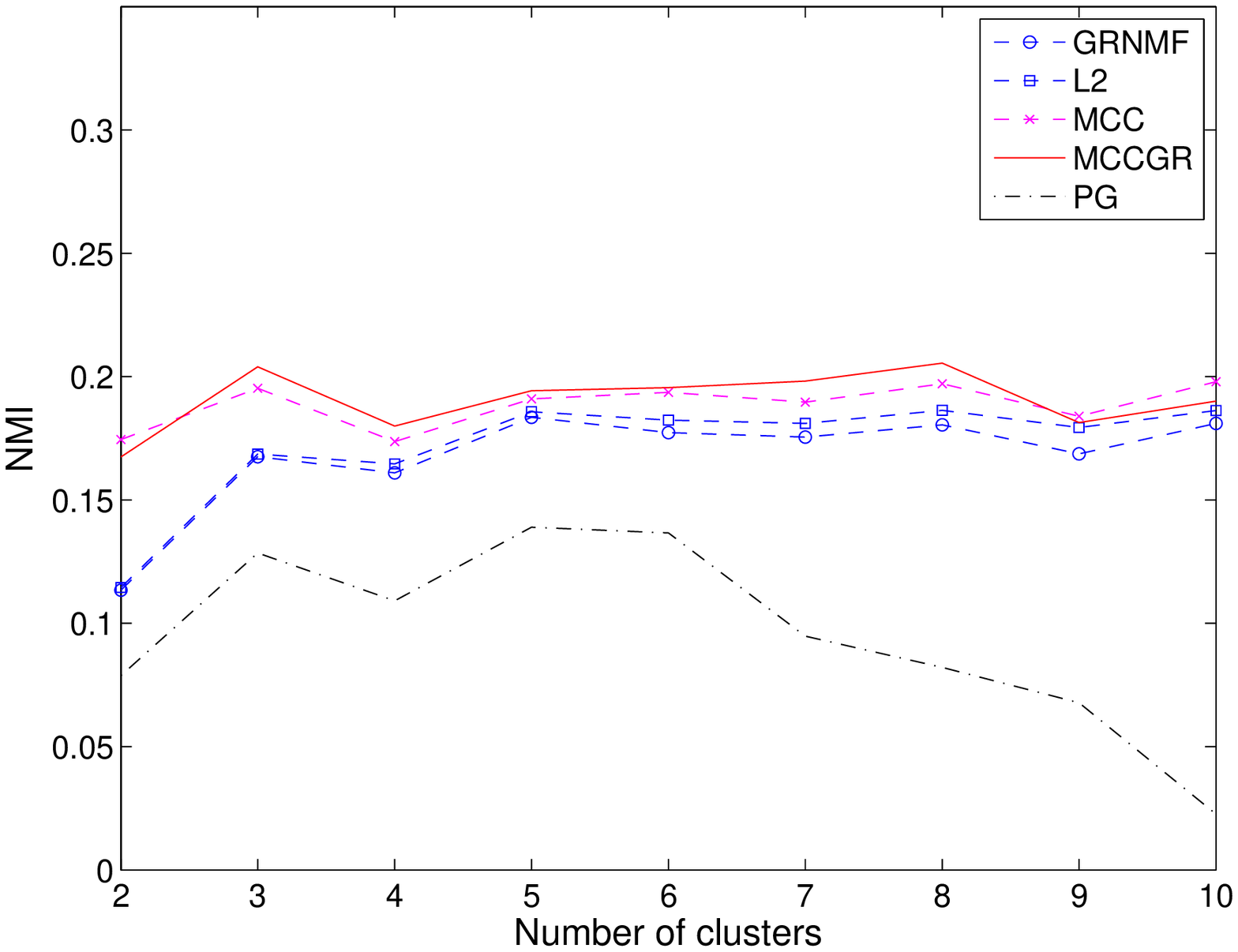}
}
\caption{MIhat on Caltech101 and Caltech256 datasets, with 300- and 1000- codewords.}
\label{fig:mihat}
\end{figure*}

\begin{table*}[ht]
\centering
\caption{Clustering NMI on Caltech101 and Caltech256 datasets, with 300 codewords in dictionary} 
\label{tab.mihat300}
\begin{tabular}{|c||c|c|c|c|c||c|c|c|c|c|}
\hline
\multirow{2}{*}{\tabincell{c}{K}} & \multicolumn{5}{|c||}{Caltech101} & \multicolumn{5}{|c|}{Caltech256}\\
\hhline{~----------}
 & GRNMF &  L2 & MCC & MCCGR & PG  & GRNMF & L2 & MCC & MCCGR & PG \\ \hline
2 & 0.247 & 0.250 & 0.232 & \bf{0.260} & 0.235   & 0.149 & 0.166 & \bf{0.185} & 0.145 & 0.142\\ \hline 
3 & 0.244 & 0.249 & 0.233 & \bf{0.259} & 0.241  & 0.162 & 0.170 & 0.167 & \bf{0.188} & 0.167\\ \hline 
4 & 0.244 & 0.266 & 0.227 & \bf{0.276} & 0.242  & 0.171 & 0.181 & 0.177 & \bf{0.209} & 0.160\\ \hline 
5 & 0.251 & 0.250 & 0.220 & \bf{0.263} & 0.224  & 0.177 & 0.181 & 0.184 & \bf{0.203} & 0.159\\ \hline 
6 & 0.269 & 0.259 & 0.232 & \bf{0.279} & 0.206  & 0.187 & 0.189 & 0.190 & \bf{0.206} & 0.163  \\ \hline 
7 & 0.272 & 0.268 & 0.248 & \bf{0.281} & 0.215  & 0.171 & 0.178 & 0.174 & \bf{0.194} & 0.147  \\ \hline 
8 & 0.267 & 0.261 & 0.241 & \bf{0.285} & 0.220   & 0.164 & 0.177 & 0.170 &\bf{0.190} & 0.153 \\ \hline 
9 & 0.267 & 0.259 & 0.255 & \bf{0.275} & 0.223   & 0.175 & 0.177 & 0.181 &\bf{0.188} & 0.155  \\ \hline 
10 & 0.267 & 0.262 & 0.251 &\bf{0.271} & 0.205  & 0.173 & 0.181 & \bf{0.189} & 0.186 & 0.134  \\ \hline 
\end{tabular}
\end{table*}

\begin{table*}[ht]
\centering
\caption{Clustering NMI on Caltech101 and Caltech256 datasets, with 1000 codewords in dictionary} \label{tab.mihat1000}
\begin{tabular}{|c||c|c|c|c|c||c|c|c|c|c|}
\hline
\multirow{2}{*}{\tabincell{c}{K}} & \multicolumn{5}{|c||}{Caltech101} & \multicolumn{5}{|c|}{Caltech256}\\
\hhline{~----------}
 & GRNMF &  L2 & MCC & MCCGR & PG  & GRNMF & L2 & MCC & MCCGR & PG \\ \hline
2 & 0.202 & 0.192 & 0.189 & \bf{0.228} & 0.185  & 0.113 & 0.115 & \bf{0.174} & 0.167 & 0.079 \\ \hline 
3 & 0.264 & 0.236 & 0.241 & \bf{0.285} & 0.217   & 0.168 & 0.169 & 0.195 & \bf{0.204} & 0.129\\ \hline 
4 & 0.237 & 0.224 & 0.221 & \bf{0.258} & 0.165  & 0.161 & 0.165 & 0.174 & \bf{0.180} & 0.109 \\ \hline 
5 & 0.226 & 0.224 & 0.212 & \bf{0.241} & 0.158  & 0.184 & 0.186 & 0.191 & \bf{0.194} & 0.139  \\ \hline 
6 & 0.267 & 0.265 & 0.246 & \bf{0.276} & 0.199  & 0.177 & 0.182 & 0.194 & \bf{0.196} & 0.137 \\ \hline 
7 & 0.264 & 0.263 & 0.262 & \bf{0.279} & 0.176 & 0.175 & 0.181 & 0.190 & \bf{0.198} & 0.095  \\ \hline 
8 & 0.288 & 0.291 & 0.278 & \bf{0.301} & 0.130 & 0.180 & 0.186 & 0.197 & \bf{0.205} & 0.082   \\ \hline 
9 & 0.251 & 0.259 & 0.253 & \bf{0.267} & 0.096  & 0.169 & 0.179 & \bf{0.184} & 0.181 & 0.068   \\ \hline 
10 & 0.267 & 0.269 & 0.258 & \bf{0.277} & 0.085 & 0.181 & 0.186 & \bf{0.198} & 0.190 & 0.023   \\ \hline 
\end{tabular}
\end{table*}

We compare our proposed method against 5 other NMF algorithms: the original NMF method with $l_2$ distance \cite{seung2001algorithms}, gradient descent-constrained least squares (GS) \cite{shahnaz2006document},  Projected Gradient NMF (PG) \cite{lin2007projected}, GRNMF and MCC NMF. To make a fair comparison, the initial values of $H$ and $W$ are the same for all algorithms at each run. There are many ways to construct the affinity matrix. Commonly-used ways of graph construction are 0-1 weighting, heat kernel weighting, and histogram intersection kernel weighting. As discussed in \cite{cai2011graph}, a reasonable weighting method is important to the performance of methods involving graph regularization. We use 0-1 weighting for GRNMF and MCCGR NMF algorithms due to its simplicity. We believe other weighting methods may outperform this one. However, the comparison between weighting schemas is beyond the scope of this paper. 

The comparison results of accuracy for $K \le 10$ can be found in Table~\ref{tab.results300} and~\ref{tab.results1000}, and Figure~\ref{fig:accu}. The NMI results are shown in Table~\ref{tab.mihat300} and~\ref{tab.mihat1000}, and Figure~\ref{fig:mihat}. As to the accuracy, the best value of each $K$ is bold. From the two tables we can see the supremacy of our proposed method. MCCGR outperforms the rest algorithms in most cases when $K \le 10$. As to the cases that MCCGR is not the best, its performance is second to the best. GRNMF and MCC achieve the second best  performance next to MCCGR in most cases. The performance of PG drops quickly as $K$ increases. The rest algorithms fall in between MCCGR and PG. If we look further at the results in Figure~\ref{fig:accu}, when $K>10$, an obvious observation is that the accuracy of all algorithms decreases as $K$ increases. However, MCCGR is very robust to the increment of the number of clusters. When it comes to the NMI measure, we can also find that the proposed MCCGR algorithm achieves the best performance among all.

There are several reasons we believe MCCGR is superior: 
\begin{itemize}
\item Unlike $l_2$ distance, MCCGR uses nonlinear kernel (correntropy) to measure the reconstructed errors. Similarity measurement based on linear kernel may not be suitable to capture the nonlinear manifold structure of data, like images, as discussed in \cite{sandler2011nonnegative}. 
\item Compared with other NMF algorithms that also map the features into nonlinear dataspace (e.g. the projected gradient kernel method in \cite{zafeiriou2009nonlinear}), our MCCGR learns different kernels of different data features. This adaptive process somehow helps MCCGR correctly cluster most images into the right partition. Based on the experiment results of two datasets, the usage of correntropy as similarity measure shows its advantage in  image clustering task.  
\item The graph regularizer ensures that the algorithm preserves the local invariant information during factorization. Thus, the NMF on the manifold further improves the selective ability of our proposed algorithm.
\end{itemize}

%


\section{Conclusion}
\label{sec:conclusion}
In this paper, we presented a graph regularized non-negative matrix factorization method by maximizing the correntropy between the original and the approximated matrices for unsupervised image clustering task. The MCC algorithm can properly model the reconstructed errors in low-level feature space, while the graph regularization ensures that the proposed algorithm preserves the local invariance information. We investigated the effects of different parameter selections on the accuracy, proved and demonstrated the convergence of the proposed algorithm, and tested the proposed method against other NMF algorithms on Caltech101 and Caltech256 datasets with multiple feature extraction settings. The average of the 50 repeated runs shows the benefits of using this graph-regularized nonlinear similarity measure for image representation. In future, we will investigate the possibility of our proposed method in medical instrument\cite{anderson2013non}, mechanical instrument\cite{cui2012interpreting} and other related areas \cite{li2014tolerance,li2013evaluating,yang2014fairness,yang2013broadcasting,singh2012hourly,gao2013spot,gao2012facial}.

\section*{Acknowledgement}
This work was partially supported by National Natural Science Foundation of China under Grant No.61273217, 61175011 and 61171193, the 111 project under Grant No.B08004.

\bibliographystyle{IEEEtran}
\bibliography{icip_le}

\begin{thebibliography}{10}
\providecommand{\url}[1]{#1}
\csname url@rmstyle\endcsname
\providecommand{\newblock}{\relax}
\providecommand{\bibinfo}[2]{#2}
\providecommand\BIBentrySTDinterwordspacing{\spaceskip=0pt\relax}
\providecommand\BIBentryALTinterwordstretchfactor{4}
\providecommand\BIBentryALTinterwordspacing{\spaceskip=\fontdimen2\font plus
\BIBentryALTinterwordstretchfactor\fontdimen3\font minus
  \fontdimen4\font\relax}
\providecommand\BIBforeignlanguage[2]{{%
\expandafter\ifx\csname l@#1\endcsname\relax
\typeout{** WARNING: IEEEtran.bst: No hyphenation pattern has been}%
\typeout{** loaded for the language `#1'. Using the pattern for}%
\typeout{** the default language instead.}%
\else
\language=\csname l@#1\endcsname
\fi
#2}}

\bibitem{suykens1999least}
J.~A. Suykens and J.~Vandewalle, ``Least squares support vector machine
  classifiers,'' \emph{Neural processing letters}, vol.~9, no.~3, pp. 293--300,
  1999.

\bibitem{chapelle1999support}
O.~Chapelle, P.~Haffner, and V.~N. Vapnik, ``Support vector machines for
  histogram-based image classification,'' \emph{Neural Networks, IEEE
  Transactions on}, vol.~10, no.~5, pp. 1055--1064, 1999.

\bibitem{zhou2010region}
Y.~Zhou, L.~Li, T.~Zhao, and H.~Zhang, ``Region-based high-level semantics
  extraction with cedd,'' in \emph{Network Infrastructure and Digital Content,
  2010 2nd IEEE International Conference on}.\hskip 1em plus 0.5em minus
  0.4em\relax IEEE, 2010, pp. 404--408.

\bibitem{li2010adaptive}
L.~Li, Y.~Zhou, and H.~Zhang, ``Adaptive learning of region-based plsa model
  for total scene annotation,'' in \emph{International Conference on
  Information and Multimedia Technology}, 2010.

\bibitem{FXR_iccv13}
C.~Fang, Y.~Xu, and D.~N. Rockmore, ``Unbiased metric learning: On the
  utilization of multiple datasets and web images for softening bias,'' in
  \emph{International Conference on Computer Vision}, 2013.

\bibitem{FT_eccv12}
C.~Fang and L.~Torresani, ``Measuring image distances via embedding in a
  semantic manifold,'' in \emph{European Conference on Computer Vision
  ({ECCV})}, Oct. 2012, pp. 402--415.

\bibitem{qin2013social}
Z.~Qin, C.~R. Shelton, and L.~Chai, ``Social grouping for target handover in
  multi-view video,'' in \emph{Multimedia and Expo (ICME), 2013 IEEE
  International Conference on}.\hskip 1em plus 0.5em minus 0.4em\relax IEEE,
  2013, pp. 1--6.

\bibitem{qin2012improving}
Z.~Qin and C.~R. Shelton, ``Improving multi-target tracking via social
  grouping,'' in \emph{Computer Vision and Pattern Recognition (CVPR), 2012
  IEEE Conference on}.\hskip 1em plus 0.5em minus 0.4em\relax IEEE, 2012, pp.
  1972--1978.

\bibitem{yu2010adaptive}
Z.~Yu, O.~C. Au, R.~Zou, W.~Yu, and J.~Tian, ``An adaptive unsupervised
  approach toward pixel clustering and color image segmentation,''
  \emph{Pattern Recognition}, vol.~43, no.~5, pp. 1889--1906, 2010.

\bibitem{yu2011nonparametric}
Z.~Yu, O.~C. Au, K.~Tang, and C.~Xu, ``Nonparametric density estimation on a
  graph: Learning framework, fast approximation and application in image
  segmentation,'' in \emph{Computer Vision and Pattern Recognition (CVPR), 2011
  IEEE Conference on}.\hskip 1em plus 0.5em minus 0.4em\relax IEEE, 2011, pp.
  2201--2208.

\bibitem{sun2013novel}
T.~Sun, S.~Ding, and Z.~Ren, ``Novel image recognition based on subspace and
  sift.'' \emph{Journal of Software (1796217X)}, vol.~8, no.~5, 2013.

\bibitem{song2011image}
H.~Song, X.~Li, and P.~Wang, ``Image annotation refinement using dynamic
  weighted voting based on mutual information.'' \emph{Journal of Software
  (1796217X)}, vol.~6, no.~11, 2011.

\bibitem{xu2013cross}
L.~Xu, Z.~Zhan, S.~Xu, and K.~Ye, ``Cross-layer detection of malicious
  websites,'' in \emph{Proceedings of the third ACM conference on Data and
  application security and privacy}.\hskip 1em plus 0.5em minus 0.4em\relax
  ACM, 2013, pp. 141--152.

\bibitem{zhang2011distributed}
H.~Zhang, Z.~Zhang, H.~Dai, R.~Yin, and X.~Chen, ``Distributed spectrum-aware
  clustering in cognitive radio sensor networks,'' in \emph{Global
  Telecommunications Conference (GLOBECOM 2011), 2011 IEEE}.\hskip 1em plus
  0.5em minus 0.4em\relax IEEE, 2011, pp. 1--6.

\bibitem{sun2013space}
Q.~Sun, R.~Ma, Q.~Hao, and F.~Hu, ``Space encoding based human activity
  modeling and situation perception,'' in \emph{Cognitive Methods in Situation
  Awareness and Decision Support (CogSIMA), 2013 IEEE International
  Multi-Disciplinary Conference on}.\hskip 1em plus 0.5em minus 0.4em\relax
  IEEE, 2013, pp. 183--186.

\bibitem{sun2013mobile}
Q.~Sun, F.~Hu, and Q.~Hao, ``Mobile target scenario recognition via low-cost
  pyroelectric sensing system: Toward a context-enhanced accurate
  identification,'' 2013.

\bibitem{shen2013virtual}
J.~Shen, P.-C. Su, S.-c.~S. Cheung, and J.~Zhao, ``Virtual mirror rendering
  with stationary rgb-d cameras and stored 3-d background,'' \emph{Image
  Processing, IEEE Transactions on}, vol.~22, no.~9, pp. 3433--3448, 2013.

\bibitem{shen2013layer}
J.~Shen and S.-C.~S. Cheung, ``Layer depth denoising and completion for
  structured-light rgb-d cameras,'' in \emph{Computer Vision and Pattern
  Recognition (CVPR), 2013 IEEE Conference on}.\hskip 1em plus 0.5em minus
  0.4em\relax IEEE, 2013, pp. 1187--1194.

\bibitem{wang2012scimate}
Y.~Wang, W.~Jiang, and G.~Agrawal, ``Scimate: A novel mapreduce-like framework
  for multiple scientific data formats,'' in \emph{Cluster, Cloud and Grid
  Computing (CCGrid), 2012 12th IEEE/ACM International Symposium on}.\hskip 1em
  plus 0.5em minus 0.4em\relax IEEE, 2012, pp. 443--450.

\bibitem{palmer1977hierarchical}
S.~E. Palmer, ``Hierarchical structure in perceptual representation,''
  \emph{Cognitive psychology}, vol.~9, no.~4, pp. 441--474, 1977.

\bibitem{logothetis1996visual}
N.~K. Logothetis and D.~L. Sheinberg, ``Visual object recognition,''
  \emph{Annual review of neuroscience}, vol.~19, no.~1, pp. 577--621, 1996.

\bibitem{wachsmuth1994recognition}
E.~Wachsmuth, M.~Oram, and D.~Perrett, ``Recognition of objects and their
  component parts: responses of single units in the temporal cortex of the
  macaque,'' \emph{Cerebral Cortex}, vol.~4, no.~5, pp. 509--522, 1994.

\bibitem{xu2003document}
W.~Xu, X.~Liu, and Y.~Gong, ``Document clustering based on non-negative matrix
  factorization,'' in \emph{Proceedings of the 26th annual international ACM
  SIGIR conference on Research and development in informaion retrieval}.\hskip
  1em plus 0.5em minus 0.4em\relax ACM, 2003, pp. 267--273.

\bibitem{shahnaz2006document}
F.~Shahnaz, M.~W. Berry, V.~P. Pauca, and R.~J. Plemmons, ``Document clustering
  using nonnegative matrix factorization,'' \emph{Information Processing \&
  Management}, vol.~42, no.~2, pp. 373--386, 2006.

\bibitem{seung2001algorithms}
D.~Seung and L.~Lee, ``Algorithms for non-negative matrix factorization,''
  \emph{Advances in neural information processing systems}, vol.~13, pp.
  556--562, 2001.

\bibitem{sandler2011nonnegative}
R.~Sandler and M.~Lindenbaum, ``Nonnegative matrix factorization with earth
  mover's distance metric for image analysis,'' \emph{Pattern Analysis and
  Machine Intelligence, IEEE Transactions on}, vol.~33, no.~8, pp. 1590--1602,
  2011.

\bibitem{xu2004document}
W.~Xu and Y.~Gong, ``Document clustering by concept factorization,'' in
  \emph{Proceedings of the 27th annual international ACM SIGIR conference on
  Research and development in information retrieval}.\hskip 1em plus 0.5em
  minus 0.4em\relax ACM, 2004, pp. 202--209.

\bibitem{zhang2006non}
D.~Zhang, Z.-H. Zhou, and S.~Chen, ``Non-negative matrix factorization on
  kernels,'' in \emph{PRICAI 2006: Trends in Artificial Intelligence}.\hskip
  1em plus 0.5em minus 0.4em\relax Springer, 2006, pp. 404--412.

\bibitem{wang2013non}
J.~J.-Y. Wang, X.~Wang, and X.~Gao, ``Non-negative matrix factorization by
  maximizing correntropy for cancer clustering,'' \emph{BMC bioinformatics},
  vol.~14, no.~1, p. 107, 2013.

\bibitem{he2011maximum}
R.~He, W.-S. Zheng, and B.-G. Hu, ``Maximum correntropy criterion for robust
  face recognition,'' \emph{Pattern Analysis and Machine Intelligence, IEEE
  Transactions on}, vol.~33, no.~8, pp. 1561--1576, 2011.

\bibitem{chang2011software}
R.~Chang, X.~Mu, and L.~Zhang, ``Software defect prediction using non-negative
  matrix factorization.'' \emph{Journal of Software (1796217X)}, vol.~6,
  no.~11, 2011.

\bibitem{cai2011graph}
D.~Cai, X.~He, J.~Han, and T.~S. Huang, ``Graph regularized nonnegative matrix
  factorization for data representation,'' \emph{Pattern Analysis and Machine
  Intelligence, IEEE Transactions on}, vol.~33, no.~8, pp. 1548--1560, 2011.

\bibitem{cai2008non}
D.~Cai, X.~He, X.~Wu, and J.~Han, ``Non-negative matrix factorization on
  manifold,'' in \emph{Data Mining, 2008. ICDM'08. Eighth IEEE International
  Conference on}.\hskip 1em plus 0.5em minus 0.4em\relax IEEE, 2008, pp.
  63--72.

\bibitem{shen2010non}
B.~Shen and L.~Si, ``Non-negative matrix factorization clustering on multiple
  manifolds.'' in \emph{AAAI}, 2010.

\bibitem{baker1998distributional}
L.~D. Baker and A.~K. McCallum, ``Distributional clustering of words for text
  classification,'' in \emph{Proceedings of the 21st annual international ACM
  SIGIR conference on Research and development in information retrieval}.\hskip
  1em plus 0.5em minus 0.4em\relax ACM, 1998, pp. 96--103.

\bibitem{liu2002document}
X.~Liu, Y.~Gong, W.~Xu, and S.~Zhu, ``Document clustering with cluster
  refinement and model selection capabilities,'' in \emph{Proceedings of the
  25th annual international ACM SIGIR conference on Research and development in
  information retrieval}.\hskip 1em plus 0.5em minus 0.4em\relax ACM, 2002, pp.
  191--198.

\bibitem{deerwester1990indexing}
S.~C. Deerwester, S.~T. Dumais, T.~K. Landauer, G.~W. Furnas, and R.~A.
  Harshman, ``Indexing by latent semantic analysis,'' \emph{JASIS}, vol.~41,
  no.~6, pp. 391--407, 1990.

\bibitem{ding2006orthogonal}
C.~Ding, T.~Li, W.~Peng, and H.~Park, ``Orthogonal nonnegative matrix
  t-factorizations for clustering,'' in \emph{Proceedings of the 12th ACM
  SIGKDD international conference on Knowledge discovery and data
  mining}.\hskip 1em plus 0.5em minus 0.4em\relax ACM, 2006, pp. 126--135.

\bibitem{lin2007projected}
C.-J. Lin, ``Projected gradient methods for nonnegative matrix factorization,''
  \emph{Neural computation}, vol.~19, no.~10, pp. 2756--2779, 2007.

\bibitem{wang2012adaptive}
J.-Y. Wang, I.~Almasri, and X.~Gao, ``Adaptive graph regularized nonnegative
  matrix factorization via feature selection,'' in \emph{Pattern Recognition
  (ICPR), 2012 21st International Conference on}, 2012, pp. 963--966.

\bibitem{xiao2013class}
Y.-H. Xiao, Z.-F. Zhu, Y.~Zhao, and Y.-C. Wei, ``Class-driven non-negative
  matrix factorization for image representation,'' \emph{Journal of Computer
  Science and Technology}, vol.~28, no.~5, pp. 751--761, 2013.

\bibitem{liu2012constrained}
H.~Liu, Z.~Wu, X.~Li, D.~Cai, and T.~S. Huang, ``Constrained nonnegative matrix
  factorization for image representation,'' \emph{Pattern Analysis and Machine
  Intelligence, IEEE Transactions on}, vol.~34, no.~7, pp. 1299--1311, 2012.

\bibitem{sun2012unsupervised}
Q.~Sun, P.~Wu, Y.~Wu, M.~Guo, and J.~Lu, ``Unsupervised multi-level
  non-negative matrix factorization model: Binary data case.'' \emph{Journal of
  Information Security}, vol.~3, no.~4, 2012.

\bibitem{belkin2001laplacian}
M.~Belkin and P.~Niyogi, ``Laplacian eigenmaps and spectral techniques for
  embedding and clustering.'' in \emph{NIPS}, vol.~14, 2001, pp. 585--591.

\bibitem{niyogi2004locality}
X.~Niyogi, ``Locality preserving projections,'' in \emph{Neural information
  processing systems}, vol.~16, 2004, p. 153.

\bibitem{fei2007learning}
L.~Fei-Fei, R.~Fergus, and P.~Perona, ``Learning generative visual models from
  few training examples: An incremental bayesian approach tested on 101 object
  categories,'' \emph{Computer Vision and Image Understanding}, vol. 106,
  no.~1, pp. 59--70, 2007.

\bibitem{griffin2007caltech}
G.~Griffin, A.~Holub, and P.~Perona, ``Caltech-256 object category dataset,''
  2007.

\bibitem{gehler2009feature}
P.~Gehler and S.~Nowozin, ``On feature combination for multiclass object
  classification,'' in \emph{Computer Vision, 2009 IEEE 12th International
  Conference on}.\hskip 1em plus 0.5em minus 0.4em\relax IEEE, 2009, pp.
  221--228.

\bibitem{lowe1999object}
D.~G. Lowe, ``Object recognition from local scale-invariant features,'' in
  \emph{Computer vision, 1999. The proceedings of the seventh IEEE
  international conference on}, vol.~2.\hskip 1em plus 0.5em minus 0.4em\relax
  Ieee, 1999, pp. 1150--1157.

\bibitem{kuhn1955hungarian}
H.~W. Kuhn, ``The hungarian method for the assignment problem,'' \emph{Naval
  research logistics quarterly}, vol.~2, no. 1-2, pp. 83--97, 1955.

\bibitem{zafeiriou2009nonlinear}
S.~Zafeiriou and M.~Petrou, ``Nonlinear nonnegative component analysis,'' in
  \emph{Computer Vision and Pattern Recognition, 2009. CVPR 2009. IEEE
  Conference on}.\hskip 1em plus 0.5em minus 0.4em\relax IEEE, 2009, pp.
  2860--2865.

\bibitem{anderson2013non}
A.~Anderson, P.~K. Douglas, W.~T. Kerr, V.~S. Haynes, A.~L. Yuille, J.~Xie,
  Y.~N. Wu, J.~A. Brown, and M.~S. Cohen, ``Non-negative matrix factorization
  of multimodal mri, fmri and phenotypic data reveals differential changes in
  default mode subnetworks in adhd,'' \emph{NeuroImage}, 2013.

\bibitem{cui2012interpreting}
S.~Cui, R.~Manica, R.~F. Tabor, and D.~Y. Chan, ``Interpreting atomic force
  microscopy measurements of hydrodynamic and surface forces with nonlinear
  parametric estimation,'' \emph{Review of Scientific Instruments}, vol.~83,
  no.~10, p. 103702, 2012.

\bibitem{li2014tolerance}
L.~Li and M.~D. Smucker, ``Tolerance of effectiveness measures to relevance
  judging errors,'' in \emph{Advances in Information Retrieval}.\hskip 1em plus
  0.5em minus 0.4em\relax Springer, 2014, pp. 148--159.

\bibitem{li2013evaluating}
L.~Li, ``Evaluating information retrieval systems with multiple non-expert
  assessors,'' 2013.

\bibitem{yang2014fairness}
J.~Yang, Y.~Wang, K.~Hua, and W.~Wang, ``Fairness based dynamic channel
  allocation in wireless mesh networks,'' in \emph{Computing, Networking and
  Communications (ICNC), 2014 International Conference on}.\hskip 1em plus
  0.5em minus 0.4em\relax IEEE, 2014, pp. 556--560.

\bibitem{yang2013broadcasting}
J.~Yang and Z.~Fei, ``Broadcasting with prediction and selective forwarding in
  vehicular networks,'' \emph{International Journal of Distributed Sensor
  Networks}, 2013.

\bibitem{singh2012hourly}
R.~P. Singh, P.~X. Gao, and D.~J. Lizotte, ``On hourly home peak load
  prediction,'' in \emph{Smart Grid Communications (SmartGridComm), 2012 IEEE
  Third International Conference on}.\hskip 1em plus 0.5em minus 0.4em\relax
  IEEE, 2012, pp. 163--168.

\bibitem{gao2013spot}
P.~X. Gao and S.~Keshav, ``Spot: a smart personalized office thermal control
  system,'' in \emph{Proceedings of the fourth international conference on
  Future energy systems}.\hskip 1em plus 0.5em minus 0.4em\relax ACM, 2013, pp.
  237--246.

\bibitem{gao2012facial}
P.~X. Gao, ``Facial age estimation using clustered multi-task support vector
  regression machine,'' in \emph{Pattern Recognition (ICPR), 2012 21st
  International Conference on}.\hskip 1em plus 0.5em minus 0.4em\relax IEEE,
  2012, pp. 541--544.

\end{thebibliography}


\section*{}
\noindent {\bf Le Li} received Master of Mathematics on Computer Science degree from University of Waterloo, Canada in 2013, and Bachelor of Science degree from Beijing University of Posts and Telecommunications, China in 2010. His current research includes machine learning, information retrieval and computer vision.

\noindent {\bf Jianjun Yang} is an assistant professor in the University of North Georgia. He got his Ph.D. of Computer Science from University of Kentucky, USA in 2011. His research interests include Wireless Networks.

\noindent {\bf Kaili Zhao} is a Phd student at Beijing University of Posts and Telecommunications (BUPT). She got her bachelor degree at Hefei University of Technology. She was a visiting student at Ohio State University for 6 months from September in 2013. Now she is doing research in computer vision at Carnegie Mellon University.

\noindent {\bf Yang Xu} is currently an undergraduate student in Beijing University of Posts and Telecommunications. His research interests include image recognition and computer vision.

\noindent {\bf Honggang Zhang} is an associate professor and director of web search center at Beijing University of Posts and Telecommunications, China. He received bachelor degree from Shandong University in 1996, master and PhD degree from BUPT in 1999 and 2002 respectively. He was visiting scholar at Carnegie Mellon University in 2008. His current research interests are in the area of pattern recognition, machine learning and computer vision.

\noindent {\bf Zhuoyi Fan} is currently an undergraduate student in Huazhong University of Science and Technology, China. Her research interests include image processing and signal processing.

\end{document}